\documentclass[letterpaper]{article} 
\usepackage{ftnright}
\usepackage[preprint]{aaai2027}  
\usepackage[hyphens]{url}  
\usepackage{graphicx} 
\usepackage{natbib}  
\usepackage{caption} 
\usepackage{algorithm}
\usepackage{algorithmic}
\usepackage{amsmath} 
\usepackage{amssymb} 
\usepackage{mathtools} 
\usepackage{bm}       
\usepackage{multirow}
\usepackage{booktabs}
\usepackage{tabularx}
\usepackage{array}
\usepackage{pifont}

\usepackage{newfloat}
\usepackage{listings}
\DeclareCaptionStyle{ruled}{labelfont=normalfont,labelsep=colon,strut=off} 
\floatstyle{ruled}
\newfloat{listing}{tb}{lst}{}
\floatname{listing}{Listing}

\title{Alpha as an Efficiency Signal: Visibility-Routed RGBA Image-to-Video Generation}

\author{
    Zhe Li\textsuperscript{\rm 1,\rm 2}\footnotemark[1],
    Honghao Qiao\textsuperscript{\rm 2},
    Zhixin Xu\textsuperscript{\rm 2},
    Qijie Wang\textsuperscript{\rm 2,\rm 3}\thanks{Work done while interning at ByteDance.},
    Bo Peng\textsuperscript{\rm 2},
    Dawei Li\textsuperscript{\rm 2}\corresponding
}
\affiliations{
    \textsuperscript{\rm 1}Peking University,
    \textsuperscript{\rm 2}ByteDance,
    \textsuperscript{\rm 3}Tsinghua University
}

\begin{document}

\maketitle

\begin{abstract}
RGBA videos combine RGB appearance with an alpha channel, enabling animated assets to be applied across arbitrary backgrounds, which are heavily used in gaming industry. However, generating high-quality RGBA animations for games remains challenging for two reasons. First, most existing RGBA video datasets are dominated by photorealistic content, with limited coverage of game assets. 
Second, the traditional generate-then-matte pipelines estimate alpha only after RGB synthesis, so semi-transparent regions are often blurred by background, resulting in unstable matting outputs.
More recently, many methods have begun to model RGB and alpha jointly, but existing approaches are mostly text-conditioned, and still have unresolved issues in efficiency and quality.
To address these challenges, we introduce \textbf{GameAlpha-2.4K}, a 2.4K-clip game-style RGBA video dataset built with matte-friendly synthesis, multi-hypothesis alpha recovery, and compositing-based quality gates.
Using this dataset, we train a reference-conditioned RGBA video generator that jointly produces RGB frames and alpha mattes in a single pass. 
To improve efficiency, we propose a \textbf{visibility router} that identifies transparent tokens in an early stage and bypasses their later DiT updates, while $x_0$-lock guides them along the original flow-matching schedule toward self-predicted endpoints.
Our model obtains lower FVD than traditional two-stage pipelines, and the visibility router skips 35\% of token evaluations in the final two DiT denoising steps, providing a $1.2\times$ backbone speedup with negligible quality degradation compared to dense inference.


\end{abstract}






\section{Introduction}

We study reference-conditioned RGBA image-to-video generation. Given a single RGBA reference image and a text prompt, the model jointly generates foreground and a temporally coherent alpha matte.

Existing solutions fall into two families, each leaving part of the problem unresolved.
The first generates an RGB video and then estimates alpha with segmentation or video matting~\cite{ravi2024sam2,yang2025matanyone,lim2026videomama}.
This decomposition is often limited for semi-transparent content: once background color has mixed into a pixel, estimating alpha alone is usually insufficient to recover a clean foreground for re-compositing; hard segmentation discards partial opacity outright, and matting tends to flicker around hair, glow, and motion blur---failures that concentrate exactly in the soft regions on which compositable assets are judged.
The second generates RGBA directly and avoids the decomposition, but open reference-conditioned models remain scarce: image-level methods carry no motion, open video models are mostly text-to-video, and image-to-video attempts are domain-specific or not fully released~\cite{wang2025transpixeler,dong2025wanalpha,zhang2026transtext,zhang2024transparent,chen2026transanimate,li2025transvdm}.
How alpha is represented further sets the cost of direct generation: TransPixeler appends alpha tokens and TransText tiles alpha beside RGB, both inflating the DiT sequence, whereas Wan-Alpha folds RGB and alpha into one merged latent at no token overhead---the representation we adopt.
Yet all existing direct generators, merged-latent designs included, spend DiT computation uniformly over spatiotemporal tokens, leaving the visibility information carried by alpha unused.
RGBA compositing provides an analytic visibility signal.
Given $C=\alpha R+(1-\alpha)B$, $\frac{\partial C}{\partial R}=\alpha$; conditioned on a region's final alpha being zero, further refinement of its RGB layer cannot affect any composite.
This criterion follows from the compositing equation, but which regions will be transparent in the output remains unknown during denoising, and alpha errors remain consequential.
The resulting problem is therefore not to estimate generic visual saliency, but to predict sufficiently early which tokens will remain fully transparent.

We use this observation as an inference-time routing principle while designing the model to generate RGB and alpha jointly.
To address the lack of reference-conditioned RGBA data for game-asset animation, we construct GameAlpha-2.4K, whose alpha mattes are recovered from videos generated under controlled, matte-friendly conditions and retained through staged quality checks.
For generation, we adapt a merged RGB-A representation to the pretrained model’s native image-to-video conditioning. 
The reference RGB and alpha are fused into a first-frame latent anchor, while its white-background composite and text enter the model’s visual and textual conditioning pathways. 
A single DoRA-adapted DiT then denoises the joint RGB-A latent without introducing additional DiT tokens or an alpha-specific DiT branch.

A visibility router reads the model’s step-2 clean estimate and predicts which tokens will remain fully transparent. 
After retaining a conservative spatiotemporal margin around the predicted foreground, the router advances the routed tokens along an \(x_0\)-locked flow path toward their step-2 predicted endpoints. Subsequent sparse steps reactivate tokens near the evolving foreground boundary, while low-support or high-uncertainty samples remain on the dense path.
Because the router changes neither the denoising schedule nor the backbone parameters, it is compatible with step distillation and other orthogonal acceleration techniques.


Our contributions are:

\begin{itemize}
  \item \textbf{Task and dataset.} We study reference conditioned RGBA image-to-video generation for compositable game-asset animation and build GameAlpha-2.4K: 2,400 clips with alpha mattes are recovered under controlled, matte-friendly conditions and verified by staged quality gates.
  \item \textbf{A visibility-aware framework.} We introduce a visibility router with $x_0$-lock that predicts effectively transparent tokens from an early clean estimate and safely reduces their late-step computation using a spatiotemporal margin and a pre-routing dense fallback.
  \item \textbf{Empirical validation.}
  On GameAlpha-2.4K, our single-stage model lowers FVD versus same-backbone generate-then-matte baselines; at a $35\%$ average evaluation 2 routed fraction, reactivation-aware routing achieves a $1.2\times$ DiT-backbone speedup with near-dense metrics and the lowest FVD among compute-matched token-selection baselines.
\end{itemize}



\section{Related Work}

\subsection{RGBA and Layered Generation}
RGBA and layered generation model appearance together with explicit opacity, producing foregrounds or layers that can be composited independently.
At the image level, LayerDiffuse~\cite{zhang2024transparent} encodes transparency in a frozen diffusion latent and ART~\cite{pu2025art} extends it to multiple layers, but neither models motion.
Video methods include text-conditioned RGBA or layered generation with TransPixeler~\cite{wang2025transpixeler}, Wan-Alpha~\cite{dong2025wanalpha}, and LayerFlow~\cite{ji2025layerflow}.
Reference-conditioned or specialized I2V methods include TransText~\cite{zhang2026transtext} for glyph animation, TransVDM~\cite{li2025transvdm} for image-and-text-conditioned synthesis, and motion-guided TransAnimate~\cite{chen2026transanimate}.
TransPixeler also demonstrates CogVideoX-based I2V, but no corresponding checkpoint or I2V code is public.
Cost separates these representations: TransPixeler and TransText enlarge the token sequence or spatial canvas, increasing attention cost, whereas Wan-Alpha folds RGB and alpha into a merged latent without token overhead. 
We combine Wan-Alpha's merged latent with native I2V conditioning and use the generated alpha to route inference computation.

\begin{figure*}[t]
  \centering
  \includegraphics[width=0.98\textwidth]{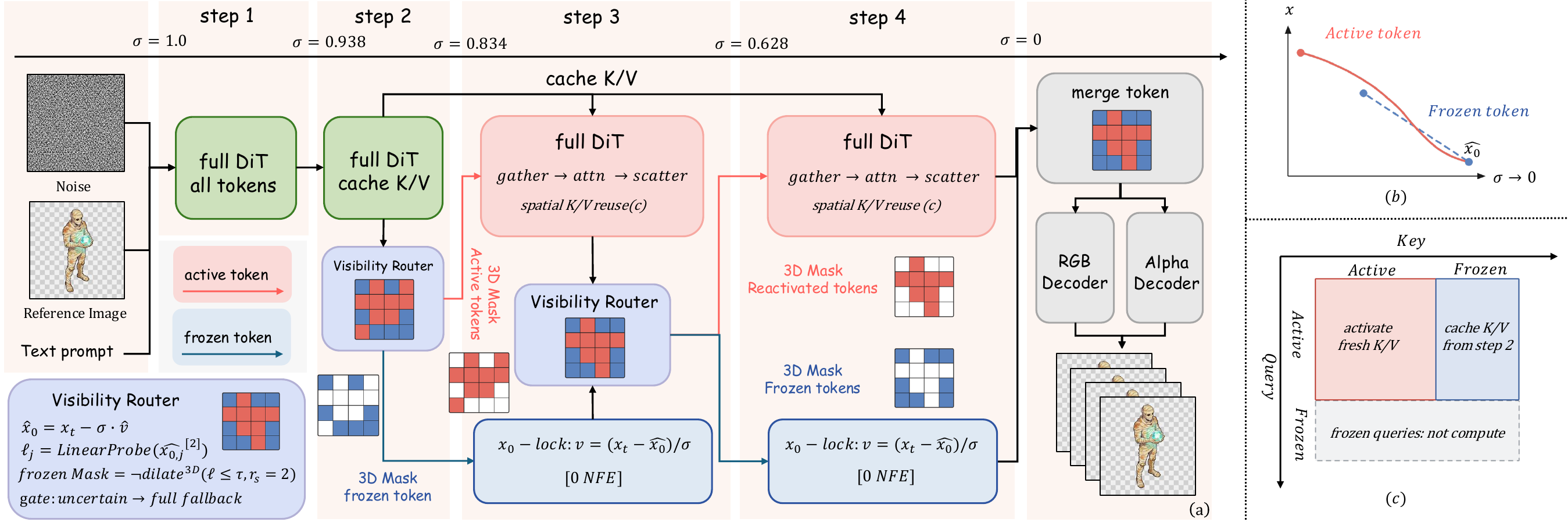}
  \vspace{-1mm}
  \caption{
  \textbf{Visibility-routed RGBA inference.}
  (a) Evaluations~1--2 process all tokens.
  After evaluation~2, the router uses $\hat{\mathbf{x}}_0^{[2]}$ to protect the predicted nontransparent support with a spatiotemporal margin and route the remaining tokens; uncertain samples retain dense inference.
  Evaluations~3--4 update active tokens, advance routed states with $x_0$-lock, and reactivate tokens reached by the evolving support before merging both sets for RGB and alpha decoding.
  (b) $x_0$-lock transports each routed state under the unchanged global noise schedule toward its stored endpoint $\mathbf{x}_{0,j}^{\star}$.
  (c) Cache-KV lets active queries attend to fresh active K/V and cached routed K/V while omitting routed queries, output projections, and feedforward updates.
  }
  \vspace{-2mm}
  \label{fig:fig_framework}
\end{figure*}

\subsection{Video Matting and Segmentation}
Two-stage pipelines generate RGB before estimating a segmentation mask or alpha matte.
SAM2/3 produce hard masks without partial opacity~\cite{ravi2024sam2,carion2025sam3segmentconcepts}, while Matting-Anything~\cite{li2023matting} estimates soft alpha at the image level.
Video matting models use temporal recurrence, memory, or global aggregation~\cite{rvm,li2022vmformer,yang2025matanyone,yang2026matanyone2}, with VideoMaMa~\cite{lim2026videomama} further introducing a generative video prior.
Despite these advances, matting remains post-hoc: a composite underdetermines foreground and alpha, so estimating alpha alone after RGB synthesis cannot guarantee a background-independent foreground layer. 
We instead generate paired foreground color and alpha in one pass.

\subsection{Efficient Diffusion Inference}

Efficient diffusion inference exploits redundancy across denoising steps, intermediate features, and tokens.
Few-step distillation reduces denoising evaluations~\cite{lightx2v,wang2024phasedconsistencymodels}, while cross-step reuse avoids repeated feature or attention computation~\cite{liu2024timestep,ma2023deepcache,zou2024accelerating,lyu2025fastercache}.
Token-level methods selectively update or merge tokens using previous-step outputs, feature similarity, or learned scores~\cite{liu2025regionadaptivesamplingdiffusiontransformers,bolya2022tome,bolya2023tomesd,rao2021dynamicvit,rao2022dynamicvit}, but these criteria are not tied to RGBA visibility.
RGBA generation provides a task-native criterion: if final alpha is zero, foreground RGB cannot affect the composite. Our router estimates this condition from the model's evolving transparency prediction.

\section{Method}

Our framework addresses three coupled challenges: constructing audited alpha supervision for synthesized RGB clips, incorporating alpha without expanding the DiT token sequence or adding an alpha-specific DiT branch, and estimating final visibility early enough to reduce later denoising computation.
We address them with GameAlpha-2.4K, a merged RGB-A latent that preserves the pretrained I2V interface, and a conservative visibility router that routes tokens predicted to be transparent in the final output away from later DiT computation, while $x_0$-lock advances their latent states.

\subsection{The GameAlpha-2.4K Dataset}
\label{sec:data}


Existing matting datasets provide limited coverage of diverse silhouettes,
equipment, effects, and partial transparency found in game assets.
A naive sequential pipeline samples prompts, synthesizes reference images and
RGB videos, applies a single matte extractor, and filters only the final outputs.
Even with category-balanced prompts, category-dependent rejection can bias the
accepted set toward simpler concepts.
Arbitrary backgrounds make foreground and alpha recovery underconstrained,
while a single extractor propagates model-specific errors.
We address these weaknesses through acceptance-aware sampling, matte-friendly
synthesis, multiple alpha hypotheses, and stage-wise audits while failures
remain repairable.
Generating RGB on arbitrary backgrounds also makes foreground and alpha recovery underconstrained, while relying on a single matte extractor transfers its systematic errors to the entire dataset.
We address these weaknesses through acceptance-aware sampling, controlled synthesis, multiple alpha hypotheses, and stage-wise audits.

\paragraph{Balancing the dataset distribution.}
To mitigate category dependent drift in the accepted set, we apply acceptance-aware tuple sampling with compatibility constraints and adaptive weights over entity, equipment, action, style, and viewpoint; the full taxonomy, prompt-expansion procedure, and weight-update rules are provided in A.1.

\paragraph{Constraining foreground recovery before matting.}
Because a matte extractor cannot resolve foreground--background ambiguities already baked into RGB, we constrain generation before matting.
A fixed high-capacity text-to-image model synthesizes a $1920{\times}1920$ reference image with a single frame-safe subject on a content-free background.
A VLM then checks the sampled attributes, framing, background contamination, and external cast shadows.
Each rejection becomes a prompt repair instruction, and the reference is regenerated for up to three attempts.
BiRefNet extracts an alpha matte from the accepted reference image to form an RGBA reference.
The accepted reference RGB also conditions a fixed high-capacity image-to-video generator.
A raw-video gate then rejects inconsistent subjects, incorrect motion or orientation, external shadows, scattered debris, and content-bearing backgrounds before 
matting.

\paragraph{Alpha recovery and staged audits.}
No single extractor performs consistently well across hard boundaries, fine structures, motion blur, holes, and partial opacity.
We therefore generate three complete alpha hypotheses for each retained RGB clip using framewise BiRefNet, SAM3 video segmentation, and MatAnyone2 video matting.
Since BiRefNet processes each frame independently, we add a temporal stabilizer that warps neighboring-frame mattes to the current frame using optical flow.
This suppresses brief errors without averaging misaligned motion boundaries.
A fixed multimodal evaluator compares sampled-frame composites for subject completeness, background leakage, soft boundaries, and temporal stability, then selects one complete sequence per clip; clip-level selection avoids discontinuities from switching extractors across frames.
The selected alpha and generated RGB clip form the RGBA video.
A final audit alone cannot determine whether a failure came from the reference, RGB video, or matte, so each gate acts while repair remains possible.
The reference gate regenerates images, the raw video gate avoids running three extractors on defective clips, and matte selection compares complete hypotheses through their composites.
A final VLM and human audit check subject consistency, action, orientation, equipment, style, temporal stability, and foreground cleanliness.

Recurring failures and category imbalance update the next batch's compatibility rules, penalties, and sampling weights.

Each sample includes an RGBA reference and video, the selected grayscale alpha sequence, structured image and video prompts, and an audit record.

\subsection{End-to-End RGBA Generation}
\label{sec:framework}
\label{sec:principle}

\paragraph{Preserving pretrained interfaces.}
Reference-conditioned RGBA generation
must introduce alpha without disrupting the pretrained model's video token geometry, native reference conditioning, and motion prior.
Alpha representation therefore determines DiT cost.
TransPixeler doubles visual tokens from $L$ to $2L$, expanding the full attention matrix from $(L_{\mathrm{text}}+L)^2$ to $(L_{\mathrm{text}}+2L)^2$, while TransText enlarges the spatial canvas by concatenating RGB and replicated alpha.
%
Concatenating RGB and alpha along the channel dimension requires new projection layers, which behave like training from scratch and reduce generation diversity when RGBA training data are limited.
We therefore follow Wan-Alpha and use a merged RGB-A VAE that preserves Wan's native 16-channel latent.
For an RGBA image or video $U=(U_{\mathrm{rgb}},U_\alpha)$, define
\begin{equation}
  \Phi(U)
  =
  \mathcal{M}\!\left(
  \mathcal{E}(U_{\mathrm{rgb}}),
  \mathcal{E}\!\left(\operatorname{Rep}_3(U_\alpha)\right)
  \right),
  \label{eq:merged}
\end{equation}
Here, $\mathcal{E}$ is the frozen Wan-VAE encoder, $\operatorname{Rep}_3$ replicates grayscale alpha into three channels, and $\mathcal{M}$ is the learned RGB-A merge module.
The target latent $\mathbf{x}_0=\Phi(V_{\mathrm{tar}})$ retains the pretrained geometry and normalization.
Separate decoders recover RGB and alpha from the merged latent without extra DiT tokens or an alpha-specific DiT branch.

\paragraph{Reusing rather than rebuilding image conditioning.}
A compact alpha latent still requires reference control.
Controllers such as SparseCtrl, I2V-Adapter, and VACE~\cite{guo2023sparsectrl,guo2024i2v,vace} allow text-to-video backbones to accept image input, but they do not reduce the cost of an expanded alpha representation.
Thus, we reuse Wan-I2V's native first-frame latent, CLIP-image, and umT5-text conditioning~\cite{radford2021learning,chung2023unimaxfairereffectivelanguage}.
Let $\mathbf{m}$ denote Wan-I2V's native four-channel first-frame mask, active only at the first latent frame.
We form an RGBA conditioning video whose first frame is $I_{\mathrm{ref}}$ and whose subsequent frames are zero:
\begin{equation}
  \begin{aligned}
    &I_{\mathrm{cond}}
    =
    \operatorname{Concat}_{T}(I_{\mathrm{ref}},0,\ldots,0),\;
    \mathbf{z}_{\mathrm{cond}}=\Phi(I_{\mathrm{cond}}),\\
    &\mathbf{y}
    =
    \operatorname{Concat}_{C}(\mathbf{m},\mathbf{z}_{\mathrm{cond}})
    \in\mathbb{R}^{20\times T_\ell\times H_\ell\times W_\ell},\\
    &\mathbf{u}_t
    =
    \operatorname{Concat}_{C}(\mathbf{x}_t,\mathbf{y})
    \in\mathbb{R}^{36\times T_\ell\times H_\ell\times W_\ell}.
  \end{aligned}
  \label{eq:i2v_condition}
\end{equation}
The white-background reference composite and prompt produce $\mathbf{c}_{\mathrm{img}}$ and $\mathbf{c}_{\mathrm{text}}$ through the pretrained CLIP and umT5 pathways, respectively, and enter native decoupled cross-attention.
Since $\mathbf{z}_{\mathrm{cond}}$ has 16 channels, combining it with the four-channel mask gives the 20-channel condition $\mathbf{y}$; concatenating $\mathbf{y}$ with the 16-channel target $\mathbf{x}_t$ preserves the original 36-channel DiT input without extra alpha tokens or a ControlNet branch~\cite{zhang2023adding}.

\paragraph{Adapting the VAE and DiT.}
The VAE must learn a general RGB and alpha representation, whereas the DiT learns game asset motion conditioned on the reference.
Stage~1 learns the RGB-A latent contract using Wan-Alpha's VAE objective on the GameAlpha-2.4K training split and VideoMatte240K~\cite{lin2021real}.
We freeze the pretrained encoder and base decoder, optimize only $\mathcal{M}$ and LoRA adapters~\cite{hu2021lora} in the separate RGB and alpha decoders, and then freeze the complete RGB-A VAE.
With this latent contract fixed, Stage~2 adapts the pretrained Wan-I2V DiT on the GameAlpha-2.4K training split. 
Given Gaussian noise $\boldsymbol{\epsilon}\sim\mathcal{N}(\mathbf{0},\mathbf{I})$ and timestep $t$ with noise level $\sigma_t$, flow matching~\cite{lipman2022flow} defines
\begin{equation}
  \mathbf{x}_t=(1-\sigma_t)\mathbf{x}_0+\sigma_t\boldsymbol{\epsilon},\;
  \mathbf{v}=\boldsymbol{\epsilon}-\mathbf{x}_0,\;
  \hat{\mathbf{x}}_0^{(t)}
  =\mathbf{x}_t-\sigma_t\hat{\mathbf{v}}_t .
  \label{eq:fm}
\end{equation}
\begin{equation}
    \hat{\mathbf{v}}_t=
\hat{\mathbf{v}}_\theta(
\mathbf{u}_t,
\mathbf{c}_{\mathrm{img}},
\mathbf{c}_{\mathrm{text}},t)
\end{equation}
We retain the base scheduler's timestep-weighted objective
\begin{equation}
  \mathcal{L}_{\mathrm{FM}}
  =
  \mathbb{E}_{\mathbf{x}_0,\boldsymbol{\epsilon},t}
  \left[
    \omega(t)\,
    \operatorname{MSE}
    \left(
      \hat{\mathbf{v}}_t,
      \mathbf{v}
    \right)
  \right].
  \label{eq:fm_loss}
\end{equation}
where $\omega(t)$ is the scheduler's timestep-dependent weight.
We freeze the base DiT and train only DoRA modules~\cite{liu2024dora} in its attention and feedforward layers.

Although decoded RGB does not affect a composite wherever $\alpha=0$, each merged latent token jointly represents RGB and alpha, so downweighting it would also weaken zero-alpha supervision. The generator therefore retains the spatially uniform, scheduler-weighted objective in Eq.~\eqref{eq:fm_loss}, while predicted visibility is used only by the inference router. The frozen decoders output aligned RGB and grayscale alpha videos without matting.

\subsection{Visibility-Routed Inference}
\label{sec:inference}

\paragraph{Predicting final transparency.}
Generic accelerators infer importance from attention, feature similarity, or update magnitude, whereas RGBA provides an analytic criterion.
For RGB $R$, alpha $\alpha$, and background $B$, let
$C=\alpha R+(1-\alpha)B$.
For a prediction $(\hat R,\hat\alpha)$ on the same background, define
$\Delta R=\hat R-R$ and $\Delta\alpha=\hat\alpha-\alpha$.
Then
\begin{equation}
  \begin{aligned}
    \frac{\partial C_c}{\partial R_c}
    &=
    \alpha,
    \qquad c\in\{r,g,b\},\\
    \hat C-C
    &=
    \alpha\odot\Delta R
    +
    \Delta\alpha\odot(\hat R-B).
  \end{aligned}
  \label{eq:visibility}
\end{equation}
Thus, RGB has no direct compositing contribution where final alpha is zero.
The criterion is analytic at the final output, but routing must predict which tokens will be transparent during denoising.

We do not use the dataset target alpha as a routing label because it may not align with the generated motion. Likewise, simply repeating the first-frame alpha is also insufficient because it cannot anticipate future motion.
We instead use the clean estimate
$\hat{\mathbf{x}}_0^{[k]}=
\mathbf{x}_k-\sigma_k\hat{\mathbf{v}}_k$
at DiT evaluation $k$, following Eq.~\eqref{eq:fm}.
Square brackets index evaluations rather than diffusion time.
A logistic-regression probe, fitted offline on dense trajectories from the training clips and frozen for sparse inference, maps each token's $2{\times}2{\times}16$ latent patch to a 64-dimensional feature and transparency logit $\ell_j$.
Dense final generations provide self-labels: token $j$ is effectively transparent if the maximum decoded alpha over its spatiotemporal region $\Omega_j$ is at most $\eta=0.05$, with larger $\ell_j$ indicating greater confidence in this event.
Routing time trades prediction risk against the number of remaining sparse evaluations.
The risk--yield analysis in Sec.~\ref{sec:routing-calibration} supports routing after evaluation~2, so we set $k_{\mathrm{route}}=2$ and retain two evaluations for sparse execution.

\paragraph{A safety margin and dense fallback.}
Because foreground support may expand after evaluation~2, routing uses a conservative spatiotemporal margin.
With $\tau=0$, tokens satisfying $\ell_j>\tau$ are predicted transparent; we dilate the complementary nontransparent support by $r_s=2$, extend it by $r_t=1$ along latent time, and always keep the first latent frame active.
Tokens outside the protected support form the candidate routed set.
The risk--yield analysis in Sec.~\ref{sec:routing-calibration} supports $r_s=2$, while $r_t=1$ provides an adjacent-time guard.
Before constructing sparse states or caches, the sample remains on the dense path if fewer than $10\%$ of tokens are candidates or more than $15\%$ are uncertain, where uncertainty is defined by $|\ell_j-\tau|<\delta$ with $\delta=2$.

\begin{table*}[t]
\centering
\setlength{\tabcolsep}{1mm}
\begin{tabular}{@{}llccccccc@{}}
\toprule
Generator & Matte & FVD$\downarrow$ & DINO-ID$\uparrow$ & Aesthetic$\uparrow$ &
Motion Sm.$\uparrow$ &
Flow-Diff$\downarrow$ & MAD\,($10^{-3}$)$\downarrow$ & Time\,(s)$\downarrow$\\
\midrule
\midrule
\multirow{4}{*}{\shortstack[l]{Wan2.1-I2V\\(4-step)}}
 & BiRefNet   & 224.3 & 0.848 & 5.396 & 0.984 & 0.598 & 160.0 & 128.4\\
 & MatAnyone2 & 225.8 & 0.846 & \textbf{5.402} & 0.985 & 0.586 & 171.3 & 149.9\\
 & SAM3       & 221.3 & \textbf{0.849} & 5.369 & 0.985 & 0.611 & 159.6 & 126.7\\
 & UniVidX    & 276.6 & 0.695 & 5.116 & 0.986 & \textbf{0.548} & 194.9 & 4212.9\\
\midrule
\multirow{4}{*}{\shortstack[l]{CogVideoX1.5-I2V}}
 & BiRefNet   & 318.8 & 0.848 & 5.223 & 0.986 & 0.609 & \textbf{92.9} & 664.5 \\
 & MatAnyone2 & 320.2 & 0.846 & 5.234 & 0.985 & 0.620 & 93.9 & 688.0 \\
 & SAM3       & 316.2 & 0.845 & 5.202 & 0.985 & 0.688 & 94.0 & 655.7 \\
 & UniVidX    & 330.7 & 0.733 & 5.011 & 0.987 & 0.595 & 138.6 & 4749.0 \\
\midrule
\multirow{4}{*}{\shortstack[l]{HunyuanVideo-I2V}}
 & BiRefNet   & 418.6 & 0.790 & 5.142 & 0.985 & 0.688 & 261.8 & 2281.1 \\
 & MatAnyone2 & 399.6 & 0.794 & 5.121 & 0.985 & 0.660 & 236.2 & 2291.7 \\
 & SAM3       & 391.8 & 0.804 & 5.124 & 0.986 & 0.658 & 242.9 & 2272.4 \\
 & UniVidX    & 470.6 & 0.515 & 4.725 & 0.986 & 0.632 & 237.8 & 6377.5 \\
\midrule
\multicolumn{2}{l}{\textbf{Ours (visibility router)}}
 & \textbf{174.4} & 0.840 & 5.314 & \textbf{0.988} & \textbf{0.548} & 108.1 & \textbf{104.8}\\
\bottomrule
\end{tabular}
\vspace{-1mm}
\caption{Two-stage generate-then-matte comparison on the GameAlpha-2.4K validation set. \textbf{Bold} marks the best value in each column; ties at displayed precision are also bolded.}
\vspace{-2mm}
\label{tab:main}
\end{table*}

\begin{figure*}[t]
    \centering
    \includegraphics[width=0.98\textwidth]{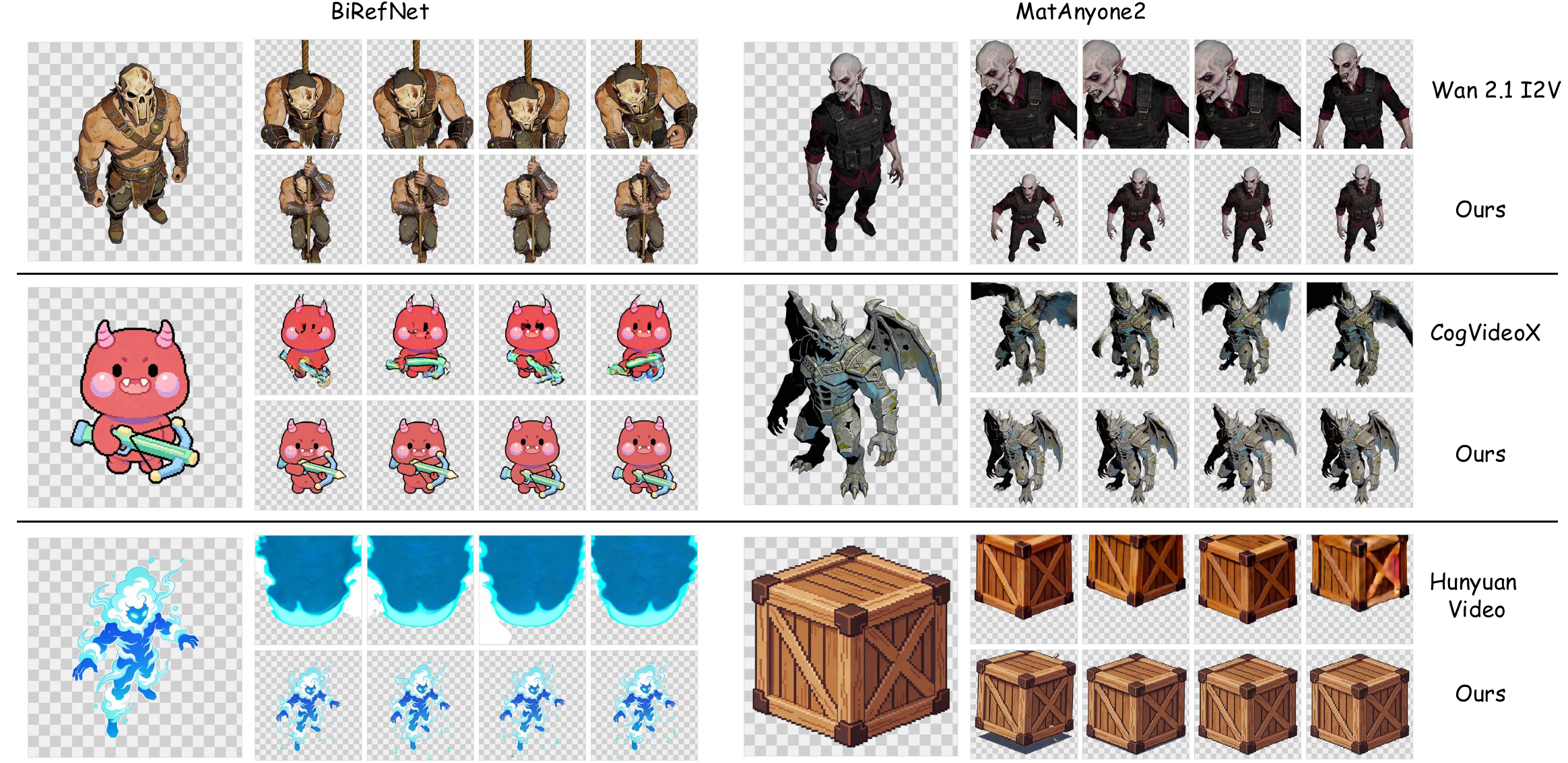}
    \vspace{-1mm}
    \caption{
    \textbf{Qualitative comparison with two-stage baselines.}
    We pair three RGB generators with BiRefNet or MatAnyone2.
    Each group shows the RGBA reference (left) and four uniformly sampled frames from the two-stage baseline (top) and ours (bottom).
    Checkerboards denote transparency; prompts are provided in the appendix D.3.
    }
    \vspace{-2mm}
    \label{fig:fig_teaser}
\end{figure*}

\paragraph{Skipping with $x_0$-lock.}
Routing determines which later token updates may be omitted, but not how the corresponding latent states should evolve.
Holding a state fixed leaves it at an obsolete noise level.
Assigning different timesteps to individual tokens creates an asynchronous schedule that conflicts with the pretrained DiT's global time conditioning.
For each token $j$ routed after $k_{\mathrm{route}}=2$, we store the model's clean endpoint
$\mathbf{x}_{0,j}^{\star}=\hat{\mathbf{x}}_{0,j}^{[2]}$.
Evaluation~2 still uses the ordinary dense scheduler update.
At each later evaluation $k>k_{\mathrm{route}}$, the routed token bypasses the DiT and follows
\begin{equation}
  \begin{aligned}
    \mathbf{v}_{k,j}^{\mathrm{lock}}
    &=
    \frac{
    \mathbf{x}_{k,j}^{\mathrm{lock}}-\mathbf{x}_{0,j}^{\star}
    }{\sigma_k},\\
    \mathbf{x}_{k+1,j}^{\mathrm{lock}}
    &=
    \mathbf{x}_{k,j}^{\mathrm{lock}}
    +
    (\sigma_{k+1}-\sigma_k)
    \mathbf{v}_{k,j}^{\mathrm{lock}}\\
    &=
    \mathbf{x}_{0,j}^{\star}
    +
    \frac{\sigma_{k+1}}{\sigma_k}
    \left(
    \mathbf{x}_{k,j}^{\mathrm{lock}}-\mathbf{x}_{0,j}^{\star}
    \right).
  \end{aligned}
  \label{eq:x0_lock}
\end{equation}
Because
$(\mathbf{x}_{k,j}^{\mathrm{lock}}-\mathbf{x}_{0,j}^{\star})/\sigma_k$
remains constant, Eq.~\eqref{eq:x0_lock} exactly transports each routed state along the straight flow-matching interpolant defined by its stored endpoint under the unchanged global noise schedule.
Only the evaluation~2 estimate is approximate.
Using token-specific endpoints also avoids discontinuities from a shared transparent latent.

\paragraph{Local reactivation and routed-token context.}
Although the evaluation-2 safety margin protects likely foreground expansion, the predicted support may still change during the sparse suffix.
After the first sparse evaluation, we update the protected nontransparent support from the fresh clean estimates of active tokens using the same spatial and temporal margins.
Any routed token reached by this updated support is reactivated for the final evaluation.
Because $x_0$-lock has advanced its state, the token re-enters the DiT at the current global noise level.
For tokens that remain routed, attention context is the remaining concern: even a token with zero final alpha may influence active queries despite having no direct RGB contribution in Eq.~\eqref{eq:visibility}.
Our primary Cache-KV implementation therefore stores the per-layer K/V of routed tokens at evaluation~2, the last dense evaluation.
During later sparse evaluations, active tokens compute fresh Q/K/V and full attention and feedforward updates, attending to fresh active and cached routed K/V; routed tokens reuse cached K/V, omit queries, output projections, and feedforward computation, and advance only through Eq.~\eqref{eq:x0_lock}.
When a token is reactivated, its cached rows are replaced with fresh K/V after re-entry.
Drop-KV removes routed context for greater speed.
Both variants retain four evaluations without changing the backbone parameters or denoising schedule.

\begin{table}[t]
\centering
{\small
\setlength{\tabcolsep}{2pt}
\begin{tabular}{@{}lrrlc@{}}
\toprule
Dataset & Clips & Frames & Domain & Public \\
\midrule
VideoMatte240K & 484 & 240.7K & human & \ding{51} \\
Wan-Alpha (DiT) & 429 & n/d & human/VFX & \ding{51} \\
TransVDM & $\sim$10K & $\sim$150K & objects & \ding{55} \\
TransAnimate & $\sim$30K & n/d & game FX/objects & \ding{55} \\
TransText & 16.1K & n/d & glyphs & \ding{55} \\
\midrule
LayerDiffuse & 1M imgs & n/a & general & \ding{55} \\
\midrule
\textbf{GameAlpha-2.4K} & 2{,}400 & 232.8K & game assets & Planned \\
\bottomrule
\end{tabular}
}
\vspace{-1mm}
\caption{Comparison of reported training data for RGBA generation or video
matting by scale, domain, and public availability. LayerDiffuse is included
as an image-only reference. ``Public'' denotes public availability at the
time of writing; n/d and n/a denote not disclosed and not applicable.}
\vspace{-2mm}
\label{tab:dataset_comparison}
\end{table}

\begin{table*}[t]
\centering
\setlength{\tabcolsep}{1mm}
\begin{tabular}{@{}lccccccc@{}}
\toprule
 & FVD$\downarrow$ & DINO-ID$\uparrow$ & Aesthetic$\uparrow$ &
Motion Smoothness$\uparrow$ & Flow Difference$\downarrow$ & MAD\,($10^{-3}$)$\downarrow$ &
Speedup$\uparrow$\\
\midrule
Dense (no routing) & 172.3 & 0.840 & 5.317 & 0.988 & 0.550 & 108.2 & 1.00$\times$\\
\midrule
\textbf{Ours (visibility router)} & \textbf{174.4} & \textbf{0.840} & \textbf{5.314} & \textbf{0.988} & 0.548 & \underline{108.1} & 1.2$\times$\\
Ours (Drop-KV)    & \underline{177.2} & \textbf{0.840} & \underline{5.304} & \underline{0.987} & 0.549 & 108.9 & 1.3$\times$\\
Ours (w/o reactivation)    & 537.2 & 0.828 & 5.209 & 0.986 & \underline{0.496} & 108.8 & 1.3$\times$\\
RAS proxy          & 819.3 & 0.823 & 5.196 & 0.985 & \textbf{0.462} & 109.5 & 1.2$\times$\\
Planner (static)   & 181.6 & \underline{0.838} & 5.290 & 0.986 & 0.547 & 108.5 & 1.2$\times$\\
Union2D (static)   & 184.2 & 0.836 & 5.273 & \underline{0.987} & 0.540 & 108.4 & 1.2$\times$\\
Random             & 297.1 & 0.820 & 5.193 & 0.984 & 0.532 & \textbf{107.7} & 1.2$\times$\\
\bottomrule
\end{tabular}
\caption{Routing ablation on the GameAlpha-2.4K validation set.
The evaluation-2 routed fraction averages $35\%$ over validation.
The no-reactivation row fixes these tokens through evaluations~3--4; Ours and Drop-KV may reactivate them and thus realize fewer skips.
RAS, Planner, Union2D, and Random match the primary router's validation-mean realized skip count.
Speedup is DiT-backbone speedup over dense inference.
\textbf{Bold}/\underline{underlined} mark best/second-best sparse quality values.}
\vspace{-2mm}
\label{tab:ablrouting}
\end{table*}

\section{Experiment}

\subsection{Setup}

\paragraph{Implementation.}
We build on Wan2.1-I2V-14B with the RGB-A VAE in Sec.~\ref{sec:framework} and four-step LightX2V distillation~\cite{lightx2v}.
We optimize rank-32 DoRA modules in the DiT attention projections $(q,k,v,o)$ and feedforward layers for 10 epochs with AdamW~\cite{loshchilov2017fixing} at $1.4\times10^{-4}$ using distributed bfloat16 training.
Training completes in less than a day on a 32-GPU node.
We generate 97 frames at $624{\times}640$.

\paragraph{Protocol.}
After splitting GameAlpha-2.4K 9:1, we train the Stage~1 RGB-A VAE on VideoMatte240K and the GameAlpha training clips, and the Stage~2 DiT only on the latter.
The transparency probe is fitted on dense training trajectories and frozen.
Analysis on the full validation split selects $k_{\mathrm{route}}=2$ and $r_s=2$; all metrics use the same split and a fixed seed.
Inference follows the four-evaluation schedule and dense fallback in Sec.~\ref{sec:inference}.

\paragraph{Metrics.}
FVD~\cite{unterthiner2019fvd,skorokhodov2021stylegan,digan,Carreira2017QuoVA} uses I3D features from checkerboard composites.
DINO-ID averages DINOv2~\cite{oquab2024dinov2} similarity between generated frames and the reference, a LAION predictor scores frame aesthetics, and motion smoothness follows VBench~\cite{huang2023vbench}.
MAD~\cite{lin2021real,yang2026matanyone2} measures alpha error against the dataset sequence in $10^{-3}$.
Flow Difference adapts TransPixeler: it compares consecutive-frame Farneback flows~\cite{Farnebck2003TwoFrameME} of the white-background RGB composite and alpha, averages endpoint distance within a dilated foreground, and normalizes by the mean magnitudes of both fields to avoid favoring static videos.
Table~\ref{tab:main} reports mean end-to-end runtime per clip, including matting, whereas Table~\ref{tab:ablrouting} reports DiT-backbone speedup over dense inference.

\subsection{Dataset Analysis}
\label{sec:exp-dataset}

At $960{\times}960$, GameAlpha-2.4K provides 97\% as many frames as VideoMatte240K across nearly five times as many clips.
Prior corpora focus on humans/VFX, generic objects, game effects, or glyphs, while LayerDiffuse is image-only; GameAlpha instead targets reusable game assets, contributing domain coverage rather than raw scale~\cite{lin2021real}.
The table does not assess matte accuracy: intermediate-opacity pixels ($2/255<\alpha<253/255$) comprise 2.2\% of GameAlpha, an opacity-distribution statistic that motivates the broader Stage~1 mixture in Sec.~\ref{sec:framework}.

\subsection{Comparison with Two-Stage Pipelines}
\label{sec:exp-baseline}


Because no reproducible open direct baseline covers this setting, we pair three I2V generators---Wan2.1-I2V-14B~\cite{wan2025}, CogVideoX1.5-I2V~\cite{yang2024cogvideox}, and HunyuanVideo-I2V~\cite{kong2024hunyuanvideo}---with four alpha extractors: BiRefNet~\cite{zheng2024birefnet}, MatAnyone2~\cite{yang2026matanyone2}, SAM3~\cite{carion2025sam3segmentconcepts}, and UniVidX~\cite{chen2026unividx}.
The four-step Wan system is closest in backbone and step budget, but RGB-A adaptation and routing make ours a full-system comparison rather than an isolated alpha ablation.
All video generators receive the same $624{\times}640$ white-composited reference and output 97 frames; ours additionally uses the RGBA latent anchor in Sec.~\ref{sec:framework}, while MatAnyone2 and SAM3 are initialized with the same reference alpha.
All systems are evaluated on the same validation videos and dataset alpha sequences.

Table~\ref{tab:main} shows that our full model achieves the lowest FVD ($174.4$) and runtime ($104.8$s).
Against the closest Wan systems, it reduces FVD and runtime.
Alpha metrics must be interpreted with content quality: our MAD is below every Wan and HunyuanVideo pipeline, although three CogVideoX combinations obtain lower MAD with much worse FVD.
Our model also achieves the best motion smoothness and keeps DINO-ID within $0.009$ of the best pipeline.
Figure~\ref{fig:fig_teaser} illustrates qualitative differences in prompt following, foreground preservation, and checkerboard composites in the displayed examples.
Overall, our full system offers the strongest observed quality--efficiency trade-off while remaining competitive in motion and RGB--alpha alignment.


\subsection{Ablation Studies}
\label{sec:exp-ablation}

Table~\ref{tab:ablrouting} compares visibility routing with a RAS-style update-magnitude proxy~\cite{liu2025regionadaptivesamplingdiffusiontransformers}, two static first-frame policies, and random selection.
Our primary router uses Cache-KV; Drop-KV retains visibility-based selection and reactivation but discards routed K/V context.
The no-reactivation variant otherwise matches the primary router but fixes the evaluation-2 routed set, isolating the value of revising early decisions.
\emph{(i) The signal matters.} Random and RAS selection can freeze visible content, degrading FVD to $297.1$ and $819.3$, respectively.
A small update at the routing evaluation does not imply final transparency, so update-magnitude saliency may freeze still-evolving content.
\emph{(ii) Adaptivity matters.} Static first-frame policies reach FVD $181.6$/$184.2$ but trail the router in DINO-ID, aesthetics, and motion smoothness, showing that fixed support cannot track evolving motion.
\emph{(iii) Visibility routing remains close to dense inference.}
Cache-KV changes FVD from $172.3$ to $174.4$ at a $1.2\times$ backbone speedup while matching dense DINO-ID and motion smoothness.
Drop-KV reaches $1.3\times$ speedup with FVD $177.2$, indicating a modest benefit from retaining routed K/V context.
MAD remains near dense across sparse variants, but lower Flow Difference need not imply better content: RAS scores $0.462$ despite an FVD of $819.3$.
Flow Difference therefore diagnoses RGB--alpha motion alignment and must be interpreted with FVD.
\emph{(iv) Reactivation corrects consequential early errors.}
Keeping the evaluation-2 routed set fixed raises FVD from $174.4$ to $537.2$ and lowers DINO-ID from $0.840$ to $0.828$, while increasing backbone speedup from $1.2\times$ to $1.3\times$.
Local reactivation therefore trades part of the nominal speed gain for recovering tokens reached by the evolving foreground.

\begin{figure}[t]
  \centering
  \includegraphics[width=0.99\linewidth]{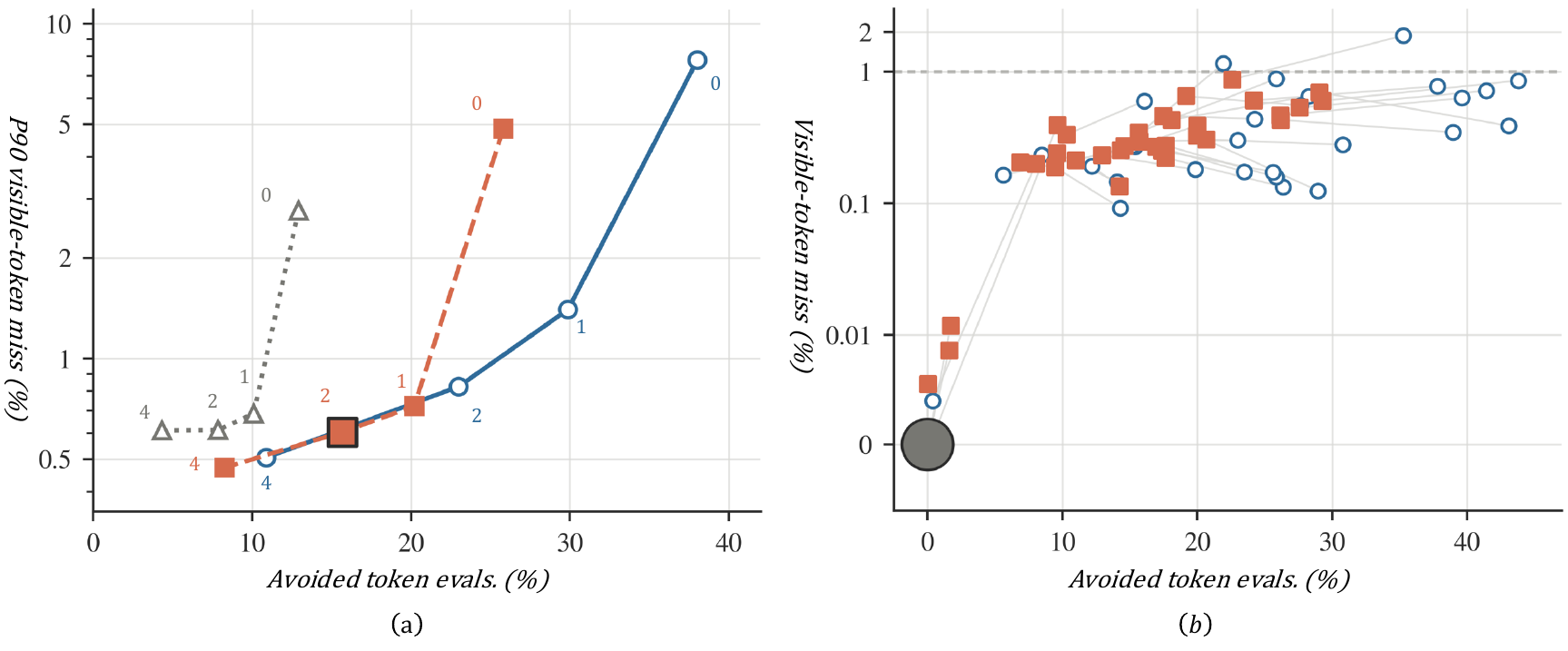}
  \vspace{-1mm}
  \caption{
  \textbf{Visibility-routing risk--yield trade-off.}
(a) The 90th-percentile final-visible-token miss rate versus the median fraction of token evaluations skipped across all four evaluations.
Blue circles, orange squares, and gray triangles indicate routing after evaluations~1, 2, and 3, respectively; labels indicate $r_s$, and the black outline identifies the selected setting, $(k_{\mathrm{route}},r_s)=(2,2)$.
(b) Per-clip changes when routing is delayed from evaluation~1 to 2 at $r_s=2$; lines connect the same clip, the gray marker summarizes clips with no savings, and the dashed line indicates a $1\%$ miss rate.}
  \vspace{-3mm}
  \label{fig:risk_yield}
\end{figure}

\subsection{Routing Time and Safety Margin}
\label{sec:routing-calibration}

Because final transparency is known only after sampling, routing must balance computation savings against the risk of omitting updates for tokens that later become visible.
Figure~\ref{fig:risk_yield} plots P90 final-visible-token miss against median full-trajectory token-evaluation saving across routing times and spatial margins.
At $r_s=2$, delaying routing from evaluation~1 to 2 reduces P90 miss from $0.824\%$ to $0.603\%$, while saving decreases from $22.99\%$ to $15.68\%$.
The $15.68\%$ value is a per-sample median over all four evaluations, whereas the $35\%$ in Table~\ref{tab:ablrouting} is the validation-set mean initial routed fraction for the final two evaluations; they differ in both denominator and aggregation.
We select evaluation~2 because it lowers tail risk while retaining two sparse evaluations.

\section{Conclusions, Limitations and Future Work}


The central observation is that alpha determines both how an RGBA asset is composited and where further RGB refinement can affect the result.
GameAlpha-2.4K supplies game-asset data, while a merged RGB-A latent lets one image-conditioned DiT generate color and alpha jointly.
This native alpha enables the visibility router to identify final-transparent tokens; $x_0$-lock advances them on the unchanged flow schedule, while conservative fallback protects uncertain clips.
Experiments show lower FVD than the evaluated generate-then-matte systems and measured quality close to dense at a $1.2\times$ DiT-backbone speedup.
The binary router targets tokens predicted effectively transparent ($\alpha\le0.05$), limiting acceleration on foreground-dense clips.
Routing errors also remain nonzero.
GameAlpha-2.4K uses synthesized videos and extractor-selected alpha, with only 2.2\% intermediate-opacity pixels.
Future work should allocate computation by expected compositing error, expand artist- or renderer-derived RGBA supervision, and test recompositability and transfer across backgrounds, domains, and video models.

\appendix

\section*{Supplementary Material}
\noindent

This supplementary material provides:
(\textbf{A}) construction details, design rationale, and statistics of the GameAlpha-2.4K dataset;
(\textbf{B}) implementation and evaluation-protocol details;
(\textbf{C}) additional experiments, analyses of the visibility router, and human evaluation;
(\textbf{D}) additional qualitative results and the prompts used by the qualitative figures;
(\textbf{E}) reproducibility notes; and
(\textbf{F}) licensing and planned release status.

\appendix

\section{The GameAlpha-2.4K Dataset}
\label{app:dataset}

\subsection{Construction Pipeline}
\label{app:pipeline}

\begin{figure*}[t]
  \centering
  \includegraphics[width=0.97\textwidth]{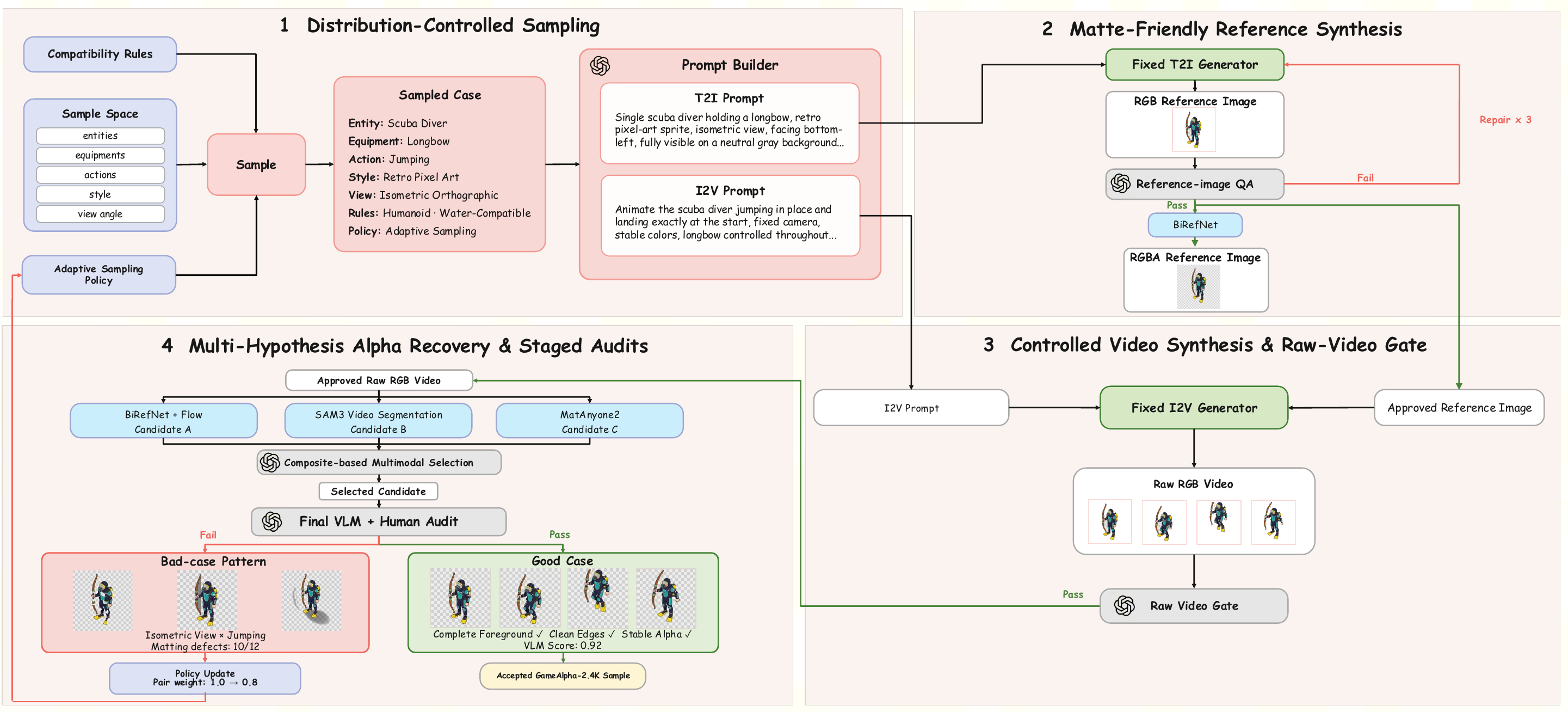}
  \caption{\textbf{The GameAlpha-2.4K construction pipeline.}
  Distribution-controlled prompt sampling (Stage 1) feeds matte-friendly reference synthesis with a repair loop (Stage 2). The approved RGB reference drives video synthesis and a raw-video gate (Stage 3), while a parallel branch forms the RGBA reference. Retained clips receive three alpha hypotheses for composite-based selection and final VLM-and-human audit (Stage 4). Recurring failures and category imbalance update the next batch’s sampling rules, penalties, and weights (red feedback edge).
  }
  \label{fig:app_pipeline}
\end{figure*}

Figure~\ref{fig:app_pipeline} summarizes the four stages of the pipeline; the
main paper describes each gate in one paragraph, and this section documents the
design rationale and the operational details that did not fit in the main text.

\paragraph{Stage 1: distribution-controlled sampling.}
Each sample starts from a tuple over entity (7 classes), equipment (7 classes),
action (5 classes), style, and viewpoint.
Hard rules remove invalid combinations, and soft penalties with category
weights suppress unlikely tuples while filling underrepresented combinations.
An LLM expands each accepted tuple into a separate image prompt and video
prompt, color metadata, and a structured description that later audits check
against.
Sampling weights are updated from the \emph{accepted} distribution and the
recorded failures rather than from the proposed distribution: acceptance rates
differ across categories, so balancing proposals alone still lets the accepted
set drift toward easy concepts.
Table~\ref{tab:app_taxonomy} lists the taxonomy.
Hard rules encode physical or perspective incompatibilities (for example,
isometric 2.5D rendering excludes strict top-down and pure side-profile
viewpoints, and object or vehicle entities exclude character-portrait
viewpoints), while soft penalties down-weight unlikely pairings such as
minimalist geometric styles with semantically strong actions.


\begin{table*}[t]
\centering
\small
\setlength{\tabcolsep}{6pt}
\renewcommand{\arraystretch}{1.12}

\begin{tabularx}{\textwidth}{
  @{}
  >{\raggedright\arraybackslash\bfseries}p{0.15\textwidth}
  >{\raggedright\arraybackslash}X
  @{}
}
\toprule
Dimension & Classes \\
\midrule

Entity (7)
& Character/NPC; Monster/Creature; Wildlife/Animal;
Vehicle/Transport; Environment Prop; Modern Tech; Food/Plant \\
\addlinespace[2pt]

Equipment (7)
& Unarmed; Melee Weapon; Ranged Weapon; Armor/Apparel;
Consumable; Tool/Gadget; Magic Relic \\
\addlinespace[2pt]

Action (5)
& Locomotion; Combat Action; State/Reaction; NPC Daily;
Physics/VFX \\
\addlinespace[2pt]

Style (23)
& Core mainstream; genre-specific; long-tail artistic.
The accepted clips cover 23 distinct styles. \\
\addlinespace[2pt]

Viewpoint (7)
& Mainstream 2D; depth/strategy; character/UI.
Seven named viewpoints are included. \\

\bottomrule
\end{tabularx}

\caption{Sampling taxonomy used to construct GameAlpha-2.4K.
Category counts are shown in parentheses. Candidate tuples span all five
dimensions; invalid combinations are filtered by hard rules, while category
weights and soft penalties shape the proposal distribution
(Appendix~\ref{app:pipeline}).}
\label{tab:app_taxonomy}
\end{table*}

\paragraph{Stage 2: matte-friendly reference synthesis.}
An extractor cannot undo foreground--background ambiguity that is already
baked into an RGB image, so the pipeline constrains generation \emph{before}
matting: a fixed high-capacity text-to-image generator renders a
$1920{\times}1920$ reference with a single frame-safe subject on a
content-free background, and a VLM gate checks the sampled attributes,
framing, background contamination, and external cast shadows.
Each rejection is converted into a prompt-repair instruction and the reference
is regenerated for up to three attempts before the tuple is returned to the
sampler as a recorded failure.
BiRefNet extracts the extractor-derived reference alpha of the accepted image
to form the RGBA reference.

\paragraph{Stage 3: controlled video synthesis and the raw-video gate.}
The accepted reference RGB conditions a fixed high-capacity image-to-video
generator.
A raw-video gate rejects inconsistent subjects, incorrect motion or
orientation, external shadows, scattered debris, and content-bearing
backgrounds \emph{before} the matting stage, so defective clips never consume
the three-extractor budget.

\paragraph{Stage 4: multi-hypothesis alpha recovery and staged audits.}
No single extractor is best across hard boundaries, fine structures, motion
blur, holes, and partial opacity, so each retained clip receives three
complete alpha hypotheses: framewise BiRefNet, SAM3 video segmentation, and
MatAnyone2 video matting.
Only the framewise BiRefNet candidate is stabilized by warping neighboring
mattes into the current frame with optical flow and taking a temporal median,
which suppresses brief errors without averaging misaligned motion boundaries;
the two video models are left untouched.
A fixed multimodal evaluator compares white-background composites of sampled
frames for subject completeness, background leakage, soft boundaries, and
temporal stability, and selects \emph{one complete sequence per clip} ---
clip-level selection avoids the discontinuities that per-frame extractor
switching would introduce.
Across the accepted corpus, the selected sequence comes from BiRefNet for
$81\%$ of clips, SAM3 for $13\%$, and MatAnyone2 for $6\%$.
The VLM audits every selected clip for subject consistency, action,
orientation, equipment, style, temporal stability, and foreground cleanliness.
Human auditors independently review a prespecified one-tenth subset as a
sampled quality-control check; this human review is not presented as a
full-corpus gate.
The main paper's shorthand ``VLM and human audit'' denotes these two distinct
coverage levels rather than a full-corpus human review.

\subsection{Gate Statistics and Audit Records}
\label{app:gates}

\begin{figure}[t]
  \centering
  \includegraphics[width=0.96\linewidth]{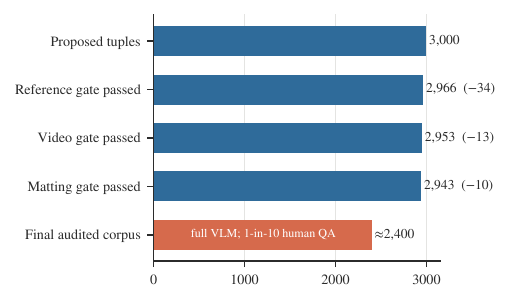}
  \caption{\textbf{Acceptance funnel of GameAlpha-2.4K.}
  From $3{,}000$ proposed tuples, $34$ are removed at the reference stage
  ($25$ rejected by the reference gate after up to three prompt-repair
  attempts, $9$ synthesis-service failures), $13$ at the video stage
  (synthesis-service and content screening), $10$ by the matte-stage quality
  gate (per-frame extraction-failure rate above $25\%$). For $19$ of the
  resulting $2{,}943$ unique tasks, production retained one additional
  regenerated clip, giving $2{,}962$ clip artifacts at the final frame-safety
  review; removing $567$
  edge-touching artifacts yields $2{,}395$ accepted clips ($2{,}156$ train /
  $239$ validation), reported as approximately $2.4$K in the main paper.}
  \label{fig:app_funnel}
\end{figure}

Figure~\ref{fig:app_funnel} reports the acceptance funnel of the accepted
batch.
Two properties are worth noting.
First, rejection is front-loaded and cheap: the reference gate and its
prompt-repair loop act before video synthesis, and the matte-stage quality
gate acts before any clip reaches the final audits, so expensive stages never
process clips that earlier stages could already reject.
Second, the final frame-safety review trades raw volume for the frame-safe
property that reference-conditioned training relies on.
Every accepted clip passes the VLM audit. Human auditors additionally review
a prespecified one-tenth subset for subject identity, action alignment,
viewpoint, equipment, and style; this sampled audit feeds the sampling-policy
iteration of Appendix~\ref{app:pipeline} but is not a full-corpus human gate.

\subsection{Category Coverage}
\label{app:coverage}

\begin{figure}[t]
  \centering
  \includegraphics[width=0.98\linewidth]{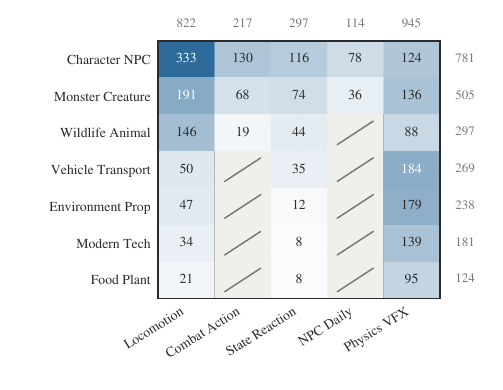}
  \caption{\textbf{Entity$\times$action coverage} of GameAlpha-2.4K.
  Hatched cells are structurally forbidden combinations under the body-type
  compatibility rules (9 of 35); every supported cell is populated (minimum
  count 8). The smallest entity class holds $5.2\%$ of clips and the
  smallest action class $4.8\%$.}
  \label{fig:app_coverage}
\end{figure}

\begin{figure}[t]
  \centering
  \includegraphics[width=0.98\linewidth]{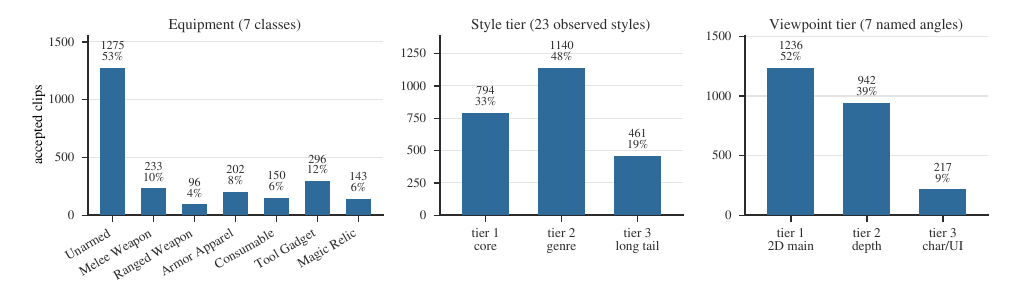}
  \caption{\textbf{Marginal distributions} over equipment, style tier, and
  viewpoint tier.
  Closed-loop reweighting keeps every category populated but not uniform:
  acceptance rates differ across categories, and sampling mixes a mainstream
  and an exploration mode, so observed shares deviate from proposal weights.}
  \label{fig:app_coverage_marg}
\end{figure}

Figures~\ref{fig:app_coverage} and~\ref{fig:app_coverage_marg} show the
realized category coverage.
The support-completeness property is the operative check for a
compositional taxonomy: all empty entity--action cells are those excluded by
hard compatibility rules (vehicles and props do not perform daily-life or
combat actions), not collapsed categories, and the smallest populated classes
retain $\approx 5\%$ of the corpus.
We report observed coverage only: proposal-side tuple distributions were not
persisted with this batch, so no quantitative before/after reweighting
comparison is claimed.

\subsection{Transparency Statistics}
\label{app:stats}

\begin{figure*}[t]
  \centering
  \includegraphics[width=0.99\textwidth]{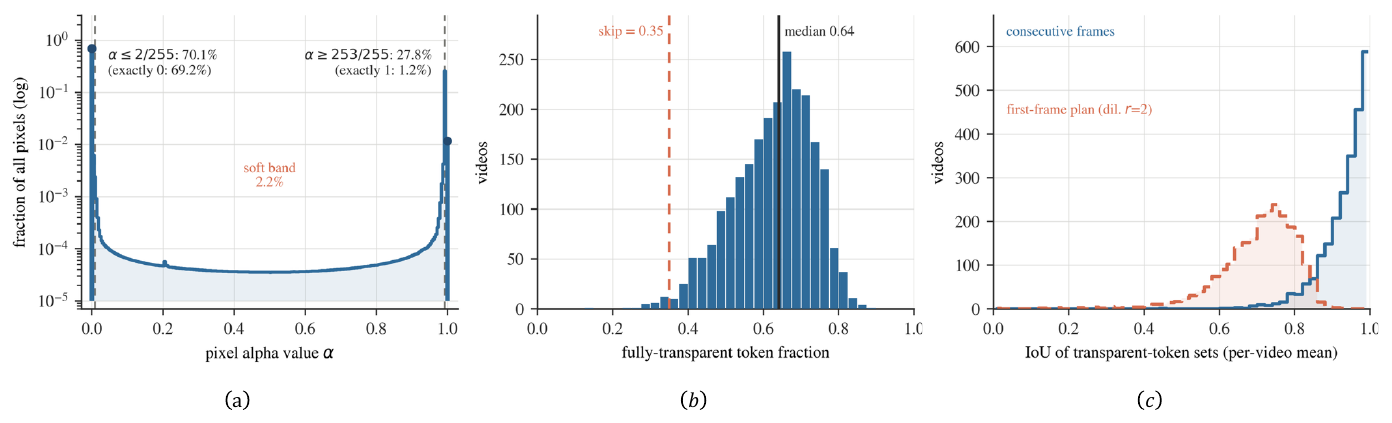}
  \caption{\textbf{Transparency structure of GameAlpha-2.4K.}
  \textbf{(a)}~Per-pixel alpha over all GameAlpha-2.4K videos $\times$ 97 frames at
  the training geometry ($624{\times}640$; log scale): $70.1\%$ of pixels are
  fully transparent ($\alpha\leq 2/255$; exactly $0$: $69.2\%$) and $27.8\%$
  effectively opaque ($\alpha\geq 253/255$; exactly $1$: $1.2\%$), leaving a
  $2.2\%$ soft band ($2/255<\alpha<253/255$); the $\pm 2/255$ margins absorb
  video-codec noise. Under the naive definition $0<\alpha<1$ the soft fraction
  is $29.6\%$, so the band definition must accompany any quoted number.
  \textbf{(b)}~Per-video fraction of fully transparent tokens under the
  conservative token criterion ($16{\times}16$ spatial and 4-frame temporal
  max-pooling, token visible if any pixel in its receptive field is): mean
  $0.63$, median $0.64$. This is the oracle upper bound on skippable
  computation; the evaluation-2 initial-routing budget for evaluations~3--4
  ($0.35$, dashed) lies below the transparent budget of $98.9\%$ of videos.
  \textbf{(c)}~Transparent-token sets of adjacent latent frames overlap almost
  perfectly (per-video mean IoU, median $0.95$), whereas a static first-frame
  plan (visible set dilated by $r{=}2$) overlaps later-frame transparency far
  less (median $0.72$): assets move and grow, and on average $12.2\%$ of
  plan-frozen tokens become visible in a later frame. This gap underlies the
  ordering of the routing ablation in the main paper.
  Token statistics here use the $2/255$ threshold of the dataset audit; the
  router's transparency labels in the main paper use $\eta=0.05$ on decoded
  alpha --- the two thresholds serve different objects and are not
  interchangeable.}
  \label{fig:app_stats}
\end{figure*}

Figure~\ref{fig:app_stats} summarizes the transparency structure of the
dataset and connects it to two design choices of the main paper.
The per-pixel distribution in panel~(a) is the domain fingerprint of RGBA game
assets: probability mass concentrates at the two ends, and
the soft band --- hair, glow, motion blur, semi-transparent effects --- is
narrow ($2.2\%$ of pixels) yet is exactly where compositing quality is judged.
This sparsity of intermediate opacity is why Stage~1 of the main paper trains
the RGB-A VAE on a broader mixture rather than on GameAlpha-2.4K alone.
Panel~(b) explains the initial routed-fraction budget: on average $62.5\%$ of tokens
are fully transparent under the conservative pooling criterion, so the
evaluation-2 $35\%$ budget for evaluations~3--4 sits well inside the transparent budget of nearly all
validation videos (and the first latent frame is never routed, which lowers
the implementation-reachable ceiling to $59.9\%$ on average).
The routed-fraction sweep in Appendix~\ref{app:sweep} confirms this headroom
empirically: Cache-KV remains within $6.7$ FVD points of the independent dense
run while realized suffix skip grows to $51.7\%$ across the tested range.
Panel~(c) quantifies why adaptivity matters: transparent sets barely move
between adjacent latent frames but drift substantially away from any static
first-frame plan, which is precisely the failure mode of the static baselines
in the routing ablation of the main paper.


\subsection{Sample Gallery}
\label{app:gallery}

\begin{figure*}[p]
  \centering
  \includegraphics[width=0.9\textwidth]{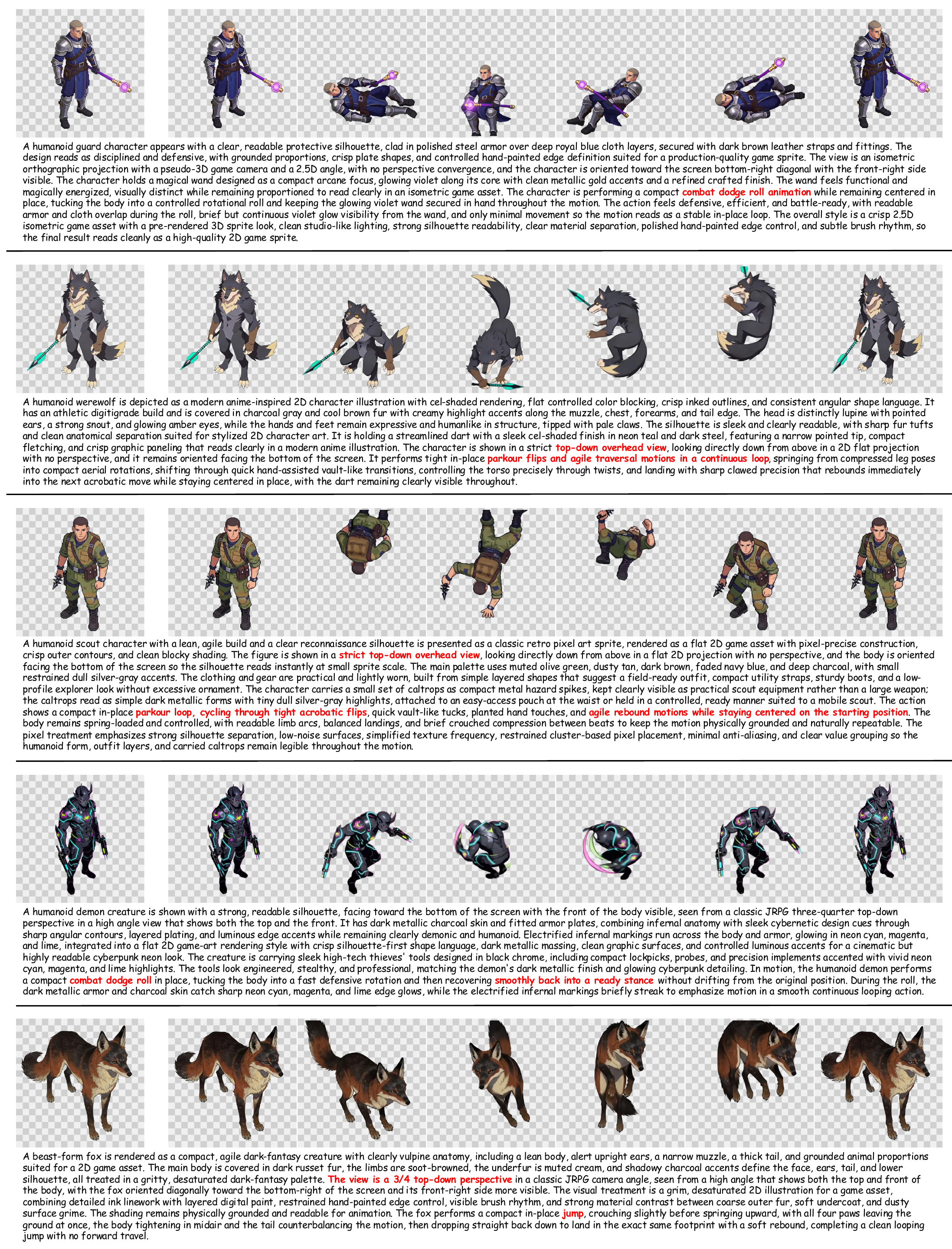}
  \caption{\textbf{GameAlpha-2.4K samples with their prompts.}
  Each row shows the RGBA reference (left), the video prompt (below;
  viewpoint and action constraints highlighted), and uniformly sampled frames
  of the clip (right); checkerboards denote transparency.
  The prompts are the structured output of the tuple sampler in
  Appendix~\ref{app:pipeline}: the sampled entity, equipment, action, style,
  and viewpoint of Table~\ref{tab:app_taxonomy} are each traceable in the
  text, and every action is constrained to be in-place and loop-closing.}
  \label{fig:app_gallery}
\end{figure*}

Figure~\ref{fig:app_gallery} shows five samples spanning distinct render
treatments (isometric 2.5D, cel-shaded, top-down pixel art, neon-accented
flat 2D, painterly dark fantasy).
Two properties recur.
First, mattes remain clean across the clip at exactly the structures where
post-hoc matting degrades --- fur tufts, glow accents, thin props.
Second, each motion is prompted to return to its starting pose within the
frame-safe canvas, making the clip suitable for looped compositing workflows.
We do not infer arbitrary-background recompositability from these examples;
that transfer test remains future work, as stated in the main paper.

\subsection{Comparison with Existing Dataset}
\label{app:dataset_comparison}

The main paper compares reported scale, domain, and availability of RGBA
training dataset.
Four clarifications belong here.
First, clip count alone understates the dataset: the exact $2{,}395$-clip
corpus contains $232.3$K frames at the $960{\times}960$ native resolution;
under the main paper's rounded $2{,}400$-clip convention, this is reported as
$232.8$K frames --- about $97\%$
of the frame budget of VideoMatte240K~\cite{lin2021real} --- distributed over
nearly five times as many clip-level examples, which matters for
reference-conditioned training where diversity is consumed at the clip level.
Second, the closest dataset differ in kind, not only in size: green-screen
human footage (VideoMatte240K), human/VFX clips (Wan-Alpha), generic objects
(TransVDM), game effects (TransAnimate), and glyphs (TransText) do not cover
reusable game-asset animation with equipment, stylization, and effect-like
foregrounds; LayerDiffuse is image-only.
Third, the alpha in GameAlpha-2.4K is recovered under controlled,
matte-friendly conditions and retained only after staged gates
(Appendix~\ref{app:pipeline}), which is a different supervision contract from
post-hoc matting of arbitrary composites.
Finally, the planned release status and licensing for GameAlpha-2.4K, the
accompanying code, and the model checkpoint are documented in
Appendix~\ref{app:licenses}.

\section{Implementation and Protocol Details}
\label{app:impl}

\subsection{DiT Adaptation and Inference Configuration}
\label{app:dit}

As stated in the main paper, we adapt Wan2.1-I2V-14B~\cite{wan2025} with
rank-32 DoRA~\cite{liu2024dora} modules on the attention projections
$(q,k,v,o)$ and feedforward layers, trained for 10 epochs with
AdamW~\cite{loshchilov2017fixing} at $1.4\times 10^{-4}$ and weight decay
$0.01$ under distributed bfloat16.
The per-GPU batch size is 1 without gradient accumulation (global batch 32).
The shifted-linear flow-matching scheduler uses shift $5$,
$\sigma_{\max}=1$, and $\sigma_{\min}=0$; training uniformly samples its
1,000 nonzero levels.
Inference uses the four-evaluation LightX2V~\cite{lightx2v} schedule shown in
the main paper, $\sigma=(1.000,0.938,0.834,0.628)$ before the zero-noise
endpoint, with FlowMatch shift~5, and generates 97 frames at
$624{\times}640$.

\subsection{Visibility Router Details}
\label{app:router}

\paragraph{Probe.}
The transparency probe is a logistic regression fitted offline on dense
denoising trajectories of training clips and frozen for routed inference.
For each token, the feature flattens its $2{\times}2$ spatial patch across the
16 latent channels of $\hat{\mathbf{x}}_0^{[k]}$ into 64 dimensions; the label
marks token $j$ effectively transparent when the maximum decoded alpha over
its spatiotemporal region $\Omega_j$ is at most $\eta=0.05$ in the dense final
output.
The probe is a readout of the model's own evolving prediction, not an external
saliency heuristic, and its cost is negligible relative to a DiT evaluation
(a single $N\times 64$ matrix product per routing decision).
Measured with the final checkpoint at inference precision (medians over
repeated runs on one accelerator), reading the probe over all $39{,}000$
tokens costs $0.8$\,ms per routing decision along the inference code path
($0.05$\,ms once inputs are resident on the accelerator), decoding alpha from
$\hat{\mathbf{x}}_0$ through the VAE alpha decoder costs $6.9$\,s, and one
dense DiT evaluation costs $22.6$\,s: probe-based routing adds
${\approx}0.004\%$ of one evaluation, whereas decoder-based routing would add
${\approx}31\%$ of one evaluation per read.

\paragraph{Margins, gates, and reactivation.}
With $\tau=0$, tokens with logit $\ell_j>\tau$ are predicted transparent; the
complementary nontransparent support is dilated by $r_s=2$ in space and
$r_t=1$ along latent time, and the first latent frame always stays active.
Before any sparse state is constructed, a sample stays on the dense path if
fewer than $10\%$ of tokens are routing candidates or more than $15\%$ are
uncertain ($|\ell_j-\tau|<\delta$, $\delta=2$); because this gate runs before
any mutation, the fallback is bit-exact dense inference.
After the first sparse evaluation, the protected support is recomputed from
the fresh clean estimates of active tokens with the same margins, and any
routed token reached by the updated support is reactivated for the final
evaluation, re-entering at the current global noise level thanks to the
$x_0$-lock transport.

\begin{algorithm}[t]
\caption{Visibility-routed sampling (four evaluations). $g$: frozen
transparency probe; margins $(r_s,r_t)=(2,1)$; the first latent frame is never
routed; with routing disabled (or on fallback) sampling is bit-exact to the
dense sampler.}
\label{alg:routing}
\begin{algorithmic}[1]
\STATE run evaluations 1--2 dense; keep $\hat{\mathbf{x}}_0^{[2]}$
\STATE $\ell \gets g(\hat{\mathbf{x}}_0^{[2]})$;\quad
$P \gets \mathrm{Dilate}_{r_s,r_t}(\{\ell_j\le\tau\}) \cup \{\text{frame }0\}$;\quad
$F \gets \complement P$
\STATE \textbf{if} $|F|/N<10\%$ \textbf{ or } $|\{j:|\ell_j-\tau|<\delta\}|/N>15\%$
\textbf{ return} dense sampling
\STATE store endpoints $\mathbf{x}^{\star}_{0,j}\gets\hat{\mathbf{x}}_{0,j}^{[2]}$
and per-layer K/V of $F$ \COMMENT{Cache-KV}
\FOR{$k=3,4$}
  \STATE active tokens: fresh Q/K/V, attention over fresh + cached K/V, FFN
  \STATE routed tokens: advance by the $x_0$-lock transport (main paper,
  Eq.~7) \COMMENT{no DiT pass}
  \STATE \textbf{if} $k=3$: recompute $P$ from fresh
  $\hat{\mathbf{x}}_0^{[3]}$ of active tokens; reactivate $F\cap P$ (fresh K/V
  on re-entry)
\ENDFOR
\STATE \textbf{return} RGB and alpha decoded from the merged states
\end{algorithmic}
\end{algorithm}


\subsection{Baseline Systems}
\label{app:baselines}

All two-stage baselines receive exactly the input our model receives --- the
reference foreground composited on white at $624{\times}640$ --- and produce
97 frames; the memory-based extractors (MatAnyone2~\cite{yang2026matanyone2},
SAM3~\cite{carion2025sam3segmentconcepts}) are seeded with the same reference
alpha our model is conditioned on, and reported runtime is the summed
generation-plus-matting wall clock on identical hardware.
\paragraph{Generators.}
Wan2.1-I2V-14B uses the official 480P weights with the same LightX2V CFG
step-distillation LoRA as our backbone (4 evaluations, no negative branch,
FlowMatch shift 5), generating natively at $624{\times}640\times 97$.
CogVideoX1.5-5B-I2V runs its official recipe (50 steps, dynamic guidance 6.0)
at its native $768{\times}768$ and 16\,fps --- a one-to-one temporal mapping
to the protocol --- and is then resampled to $624{\times}640$.
HunyuanVideo-I2V (13B) runs its official step count and guidance directly at
the protocol resolution $624{\times}640$ (both sides divisible by 16); its
720p bucket is computationally infeasible at 97 frames, which we disclose
rather than crop.
All generator outputs are conformed to $624{\times}640$, 97 frames, 16\,fps
with high-quality H.264 (alpha read-back error ${\approx}2\times 10^{-4}$
MAE), and all generators share seed 0.
\paragraph{Extractors.}
BiRefNet uses the RMBG-2.0 release (framewise, $1024^2$ input, soft alpha);
MatAnyone2 propagates from the first-frame dataset alpha (binarized
seed mask); SAM3 tracks from the same seed and produces binary masks (largest
connected component); UniVidX generates alpha with its released configuration.
Per-clip runtime sums generation and matting wall clock, excluding model
loading.

\subsection{Metric Definitions}
\label{app:metrics}

\paragraph{FVD.}
I3D features~\cite{Carreira2017QuoVA} on checkerboard composites of RGBA
outputs, following~\cite{unterthiner2019fvd,skorokhodov2021stylegan}; both generated and
reference sets use the same checker pattern and compositing code.
The checker uses $16$-pixel tiles with light/dark gray levels $230/179$ (XOR
pattern, dark origin) at the evaluation resolution $624{\times}640$, identical
for every frame and every system. For the GameAlpha-2.4K evaluation, FVD uses
exactly 97 frames per clip and all $n{=}239$ validation clips (FVD values are
not comparable across different $n$ or frame counts); cross-domain cohorts are
specified separately in Appendix~\ref{app:ood}.
\paragraph{DINO-ID.}
Mean DINOv2~\cite{oquab2024dinov2} ViT-S/14 cosine similarity between 16
uniformly sampled generated frames (white composites) and the reference
frame's white composite.
\paragraph{Aesthetic.}
Mean LAION aesthetic-predictor score over 8 uniformly sampled white-composite
frames.
\paragraph{Motion smoothness.}
The VBench~\cite{huang2023vbench} motion-smoothness metric.
\paragraph{MAD.}
Mean absolute difference ($\times 10^{-3}$) between generated and dataset
alpha sequences~\cite{lin2021real,yang2026matanyone2}, over all pixels of all
97 frames at $624{\times}640$ (alpha decoded as the channel mean of the
grayscale video). For GameAlpha-2.4K, the dataset alpha is extractor-selected
supervision rather than renderer-native ground truth; consequently, MAD
measures agreement with that reference sequence and does not by itself
establish physical opacity accuracy or arbitrary-background recompositability.
\paragraph{Flow Difference.}
Following the adaptation described in the main paper: Farneback
flows~\cite{Farnebck2003TwoFrameME} are computed on consecutive frames of the
white-background RGB composite and of the alpha sequence; the endpoint
distance between the two flows is averaged inside a dilated foreground mask
and normalized by the mean magnitudes of both fields, so static videos gain no
trivial advantage.
Flow Difference diagnoses RGB--alpha motion alignment and is interpreted
jointly with FVD (a frozen-content clip can have excellent alignment).
Precisely, for each consecutive pair $(t{-}1,t)$: Farneback flows (parameters
$0.5,3,15,3,5,1.2,0$) are computed on the grayscale white composite and on
the alpha map; the foreground mask is the union
$\max(\alpha_{t-1},\alpha_t)>0.05$ dilated twice with a $5{\times}5$ kernel
($\approx 4$ pixels); the pair score is
$\mathrm{mean}_{\mathrm{fg}}\lVert F_{\mathrm{rgb}}-F_{\alpha}\rVert_2 \,/\,
(\mathrm{mean}_{\mathrm{fg}}\lVert F_{\mathrm{rgb}}\rVert +
\mathrm{mean}_{\mathrm{fg}}\lVert F_{\alpha}\rVert + 10^{-3})$;
clip scores average the 96 pairs (empty-foreground pairs are skipped), and
table entries average all 239 clips.

\section{Additional Experiments and Analyses}
\label{app:exp}

\subsection{Early Readability of Token Transparency and Routing Time}
\label{app:probe-timing}

The visibility criterion of the main paper is analytic at the final output,
but membership in the final-transparent set is unknown while denoising is
still running.
This section tests whether the model's step-wise clean estimates already
contain a cheaply readable signal of \emph{final} token transparency, and
when that signal becomes reliable enough to route computation.
From held-out denoising trajectories that pass an alpha-health gate (rejecting
degenerate alpha predictions), we score every token at each of the four
evaluations with the probe fitted on training trajectories and evaluate the
same scores against two label sets:
\textbf{Final-Transparency Prediction AUROC} against
the final decoded alpha (does the current state predict the final transparent
set?), and \textbf{Same-Step Readout AUROC} against alpha decoded from the
same-step $\hat{\mathbf{x}}_0^{[k]}$ (does the probe correctly read the
model's current transparency estimate?).
Alpha maps are max-pooled to the DiT token grid; a token is visible if any
alpha in its receptive field exceeds $\eta=0.05$.

\begin{figure}[t]
  \centering
  \includegraphics[width=0.92\linewidth]{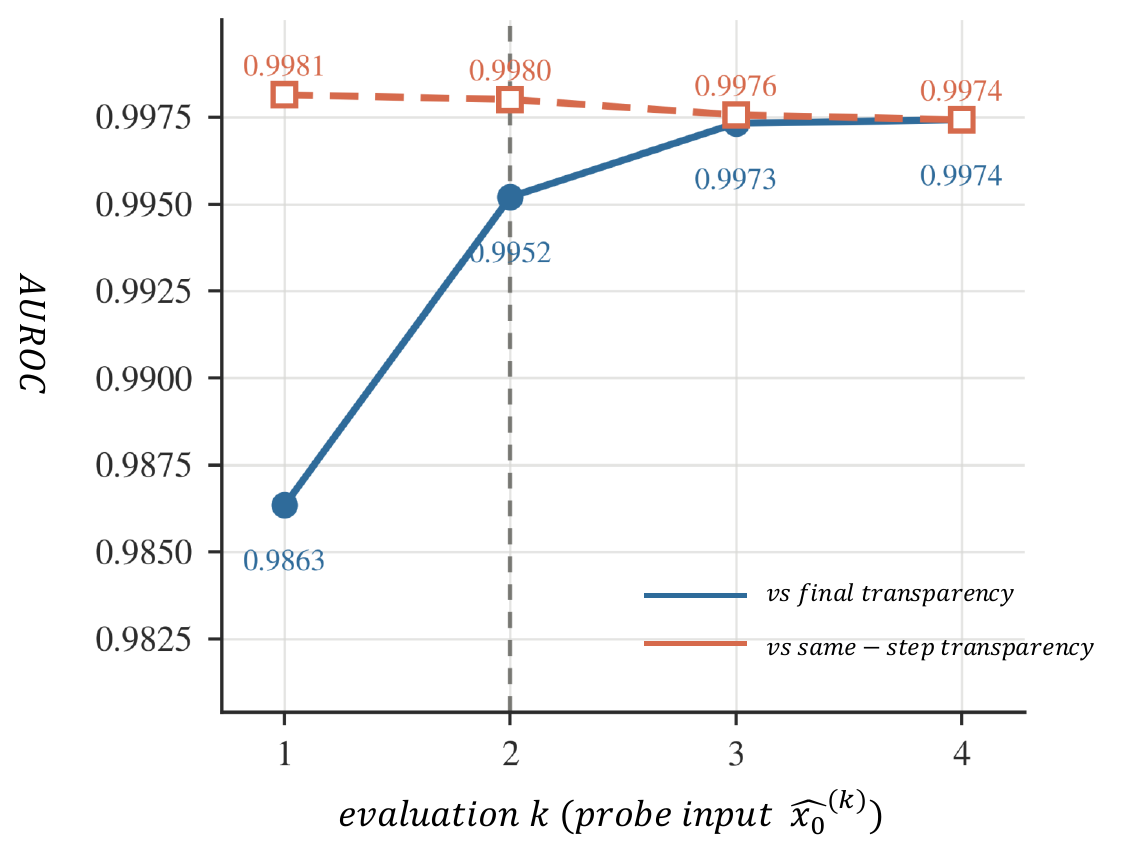}
  \caption{\textbf{Early readability of token transparency} (probe fitted on
  training trajectories;
  48 held-out validation trajectories; $1{,}872{,}000$ tokens per evaluation;
  identical token set for both curves; the alpha-health gate fired on 0 of 48
  clips).
  Blue (solid, circles): probe scores against \emph{final} transparency
  labels. Coral (dashed, squares): the same scores against \emph{same-step}
  decoded alpha. The vertical line marks the routing point $k{=}2$.
  Same-step readout stays at $0.9974$--$0.9981$ from the first evaluation,
  while final-transparency prediction rises from $0.9863$ (evaluation~1) to
  $0.9952$ (evaluation~2) and plateaus near $0.997$; at evaluation~4 the two
  targets coincide by construction. The early gap is consistent with belief
  drift of the model's own alpha estimate rather than a readout failure.}
  \label{fig:app_probe_timing}
\end{figure}

Two observations follow.
First, the probe reads the model's current transparency estimate essentially
from the first evaluation (readout $0.9981$ at evaluation~1), so early
routing is not limited by readout capacity.
Second, the current estimate itself keeps converging toward the final output:
final-transparency AUROC improves markedly from evaluation~1 to evaluation~2
($0.9863{\to}0.9952$, closing the gap to the readout curve from $0.0118$ to
$0.0028$) and only marginally afterwards ($0.9973$ at evaluation~3), which
supports $k_{\mathrm{route}}=2$ as the earliest practical routing point ---
waiting until evaluation~3 buys a small ranking gain but halves the sparse
suffix.
AUROC only selects the routing time; freezing safety is provided by the
spatiotemporal margin, local reactivation, and the pre-routing dense fallback
of the main paper, and AUROC values are ranking quality, not classification
accuracy.

\subsection{How the Alpha Trajectory Evolves}
\label{app:trajectory}

This analysis visualizes \emph{why} early routing plus a local margin is the
right shape of solution: the model's clean estimate forms the correct subject
silhouette early, and the remaining changes concentrate near the foreground
boundary as contours, soft edges, and motion details emerge.

\begin{figure*}[t]
  \centering
  \includegraphics[width=0.97\textwidth]{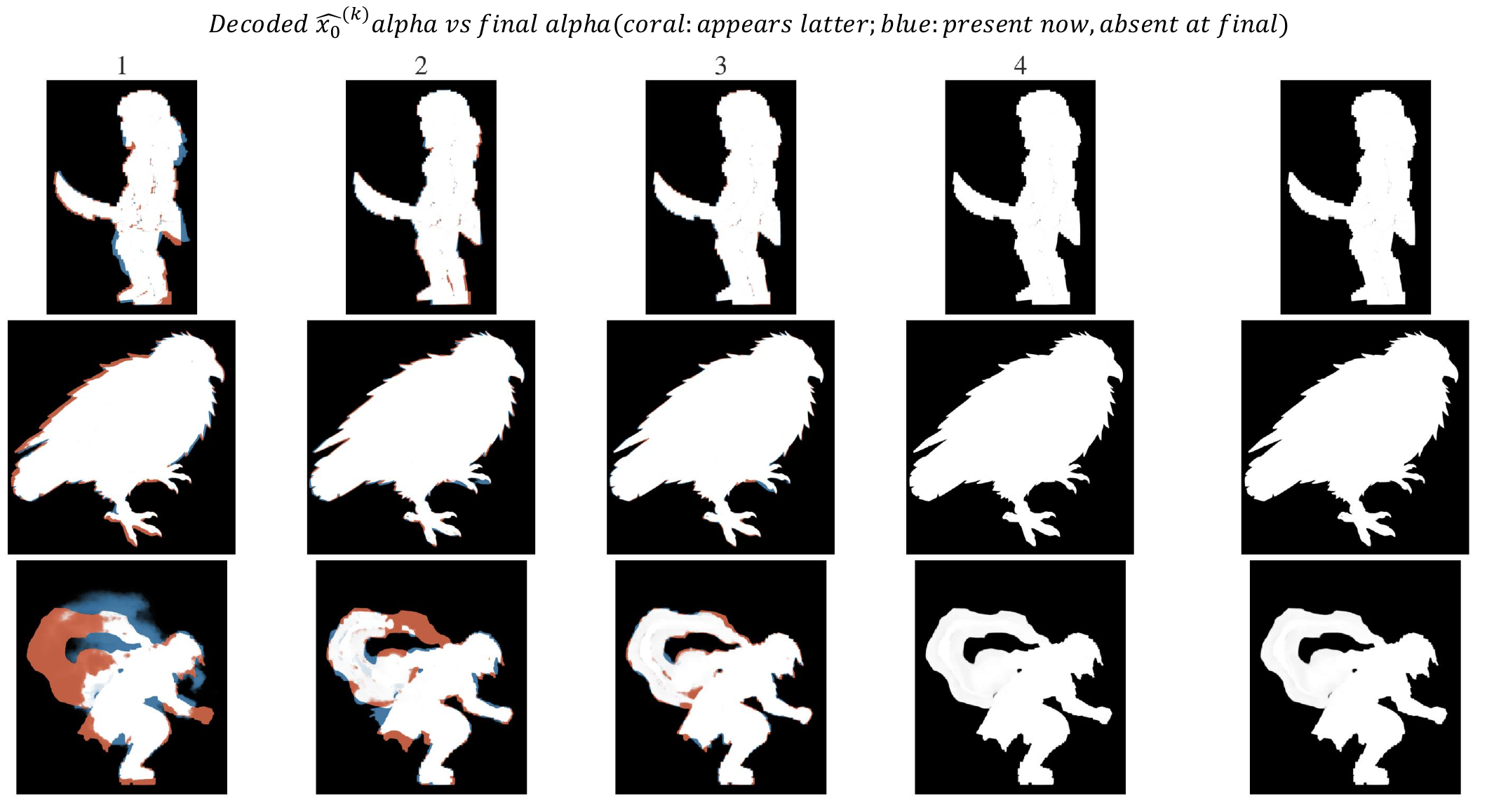}
  \caption{\textbf{Evolution of decoded alpha estimates} (deployed
  checkpoint; dense validation trajectories).
  Three manually selected alpha-healthy clips (rows) at evaluations 1--4
  (numbered columns) and the final output (rightmost column), decoded
  directly from the stored clean-estimate latents.
  Grayscale shows the current decoded alpha; coral marks support that
  \emph{appears later} (missing relative to the final alpha), and blue marks
  support \emph{present now but absent at final}.
  Rows are cropped independently, so scales are not comparable across rows.
  The visualization diagnoses trajectory self-consistency --- how the model's
  own estimate converges to its own final output --- and does not measure
  matte accuracy against an external alpha target.}
  \label{fig:app_trajectory}
\end{figure*}

Reading Figure~\ref{fig:app_trajectory} column by column: at evaluation~1 the
silhouette is already essentially correct, with residual coral/blue
disagreement forming a band around the boundary (row 3 shows the largest
early drift among the three examples); by evaluation~2 the band thins to
one or two token widths; evaluations~3--4 refine soft edges.
This boundary-concentrated convergence is exactly the pattern the router's
safety design assumes: routing after evaluation~2 with a two-token spatial
margin covers the dominant residual drift, and reactivation handles the tail
--- consistent with the risk--yield analysis in the main paper.

\subsection{Within-Frame Support-Mismatch Distances}
\label{app:mismatch}

\begin{figure}[t]
  \centering
  \includegraphics[width=0.92\linewidth]{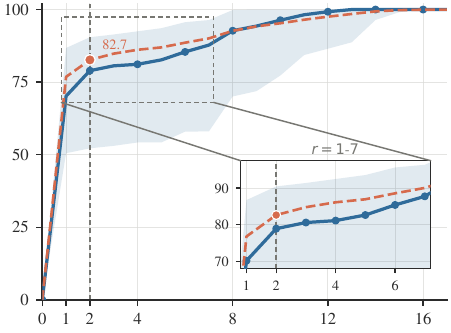}
  \caption{\textbf{Within-frame support-mismatch distance CDF (pilot).}
  For alpha-healthy analysis trajectories at $k{=}2$: cumulative fraction of
  raw per-frame support mismatches (tokens transparent in the evaluation-2
  estimate but visible in the final output) within a given 2D Chebyshev
  distance of the current foreground, \emph{before} any dilation or
  first-frame protection is applied.
  Blue: per-sample median CDF; band: P10--P90 across samples; coral dashed:
  pooled over tokens ($82.7\%$ of pooled mismatches lie within radius 2,
  marked). The inset enlarges $r=1$--$7$, where the median and pooled curves
  differ most. Per-sample mismatch counts vary widely, so the pooled curve is
  weighted toward high-mismatch samples.
  This is a mechanism diagnostic for the margin choice, not a measurement of
  the deployed 3D margin: it cannot be read as ``$r_s{=}2$ covers $82.7\%$ of
  routed misses''.}
  \label{fig:app_mismatch}
\end{figure}

Figure~\ref{fig:app_mismatch} asks where the evaluation-2 transparency
estimate disagrees with the final output within each frame: most raw
mismatches lie within one or two token cells of the current foreground
boundary, supporting the choice of a small spatial margin ($r_s=2$) over a
globally enlarged active set, while the long tail (distances up to 17) is why
the margin alone is not the entire safety story --- the reactivation pass and
the dense fallback of the main paper cover what a local margin cannot.

\subsection{Routed-Fraction Sweep}
\label{app:sweep}

\begin{figure}[t]
  \centering
  \includegraphics[width=0.96\linewidth]{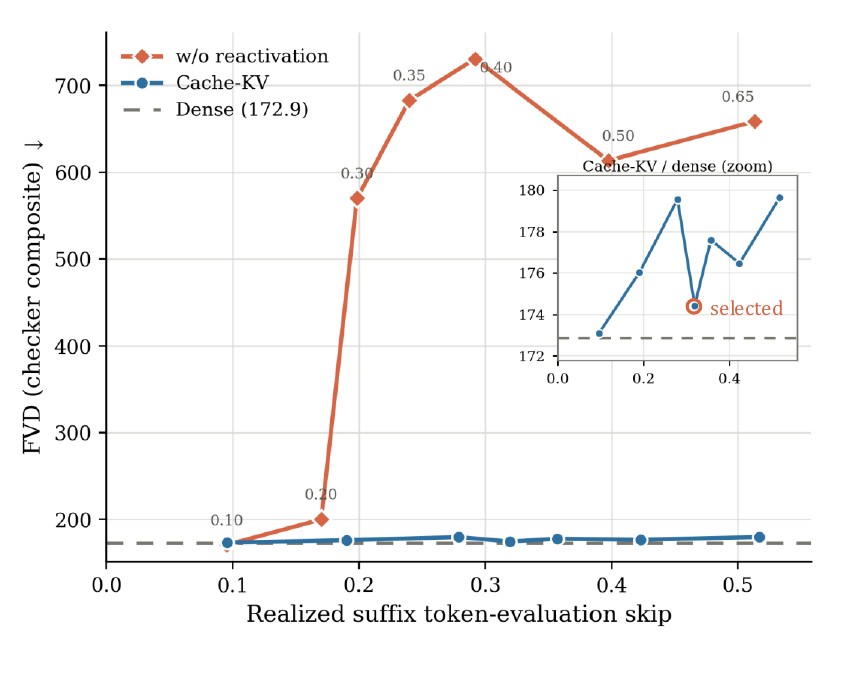}
  \caption{\textbf{Quality--budget frontier of the deployed Cache-KV router}
  (GameAlpha-2.4K validation, $n{=}239$).
  Labels denote requested routing fractions, while the horizontal axis reports
  the fraction of token evaluations actually skipped over evaluations~3--4;
  the $7/239$ dense fallbacks count as zero skip.
  Blue circles show Cache-KV with local reactivation, and the dashed line is
  dense inference under the same 97-frame checker-composite FVD protocol.}
  \label{fig:app_sweep}
\end{figure}

Figure~\ref{fig:app_sweep} sweeps the requested routing fraction over
$\{0.10,0.20,0.30,0.35,0.40,0.50,0.65\}$ under true sparse inference.
With reactivation, Cache-KV remains close to dense (FVD $173.1$--$179.6$
versus $172.9$) while the realized suffix skip grows from $9.6\%$ to
$51.7\%$.
At the requested $0.35$ tier, Cache-KV obtains FVD $174.4$ at $31.9\%$
realized suffix skip.
The controlled no-reactivation comparison is the selected operating-point row
reported in the main paper: it starts from the same evaluation-2 routed set
and dense-fallback decisions but never returns tokens to the active set,
raising FVD from $174.4$ to $537.2$ while increasing backbone speedup from
$1.2\times$ to $1.3\times$.
Under this matched construction, disabling reactivation cannot produce fewer
skips than enabling it; the no-reactivation result is therefore not mixed into
the separate budget sweep in Figure~\ref{fig:app_sweep}.
The dense value differs from the main ablation table ($172.3$) because it is
an independent run under the same metric harness.

\subsection{Fallback and Reactivation Statistics}
\label{app:fallback}

\begin{figure}[t]
  \centering
  \includegraphics[width=0.48\linewidth]{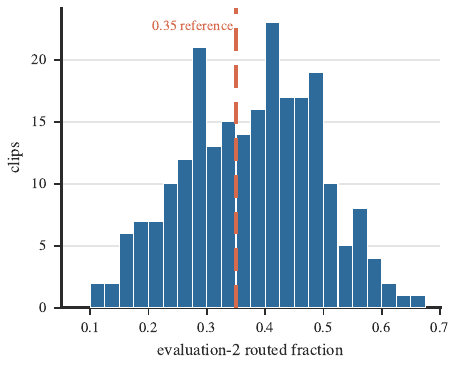}
  \hfill
  \includegraphics[width=0.48\linewidth]{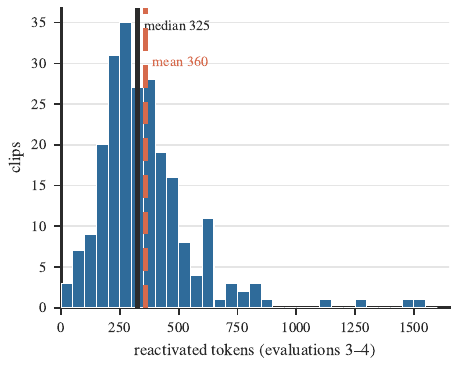}
  \caption{\textbf{Deployment statistics of the safety mechanisms}
  (validation split, $n{=}239$, primary realization).
  Left: per-clip distribution of the evaluation-2 routed fraction over the
  232 routed clips (dashed: the main paper's $0.35$ initial routed-fraction
  reference for evaluations~3--4).
  Right: per-clip reactivated-token counts between evaluations 3--4 (solid:
  median $325$; dashed: mean $360$).}
  \label{fig:app_router_stats}
\end{figure}

Figure~\ref{fig:app_router_stats} measures how often the router's safety
mechanisms actually fire on the validation split.
\emph{Dense fallback.}
The pre-routing gate keeps $7$ of $239$ clips ($2.9\%$) on the dense path:
$4$ trigger the low-support condition (fewer than $10\%$ candidates), $2$ the
high-uncertainty condition (more than $15\%$ uncertain tokens), and $1$ both.
Because the gate decides before any sparse state is constructed, these clips
execute bit-exact dense inference.
\emph{Routed fraction and full-trajectory saving.}
The $35\%$ figure in the main paper refers to the evaluation-2 initial-routing
reference for evaluations~3--4, so it describes the sparse suffix rather than
an all-step compute-saving percentage. The $15.68\%$ figure is instead the
per-sample median fraction of token evaluations avoided over evaluations
1--4, whose first two evaluations remain dense; it also reflects fallback and
reactivation. Thus the two percentages use different denominators and
aggregations rather than making competing claims about the same quantity.
\emph{Reactivation.}
Every routed clip reactivates tokens between evaluations 3--4 (median $325$
tokens, i.e., $2.1\%$ of the routed set; maximum $16.4\%$), confirming that
the evaluation-2 decision does require local revision; the cost is small,
reducing realized skip by only $0.46$ percentage points on average.

\subsection{Cross-Domain Evaluation on VideoMatte240K}
\label{app:ood}

This section evaluates cross-domain behavior on
VideoMatte240K~\cite{lin2021real}: human-subject clips outside the game-asset
training domain of our Stage-2 generator.
For this cross-domain diagnostic, we use a separately trained RGB-A VAE
checkpoint that was trained without VideoMatte240K; none of the evaluated
VideoMatte240K clips or alpha sequences was used to train that VAE, the
Stage-2 DiT, or the transparency probe.
For each clip, we take the first 97 source frames without temporal resampling
and apply to foreground and alpha a $421{\times}432$ crop centered on the
union of ground-truth alpha support ($\alpha>2/255$) over those frames.
We then resize both to $624{\times}640$ and encode them at 16 fps.
All systems receive the resulting white-composited first frame and the same
per-clip caption from a fixed open captioning model; MatAnyone2 and SAM3 also
receive the same first-frame ground-truth alpha.
Our model is evaluated with dense inference and with the visibility router at
its deployed configuration.
Metrics follow the main paper, with MAD computed against the dataset's ground
truth alpha.

\begin{table*}[t]\centering\footnotesize
\setlength{\tabcolsep}{4.5pt}
\begin{tabular}{ll cccc cc}
\toprule
Generator & Matte & FVD$\downarrow$ & DINO-ID$\uparrow$ & Aesthetic$\uparrow$ &
Motion Smoothness$\uparrow$ & Flow-Diff$\downarrow$ & MAD\,($10^{-3}$)$\downarrow$\\
\midrule
\multirow{4}{*}{\shortstack[l]{Wan2.1-I2V\\(4-step)}}
 & BiRefNet   & 133.9 & 0.887 & 5.166 & 0.991 & 0.624 & 135.9\\
 & MatAnyone2 & 132.6 & 0.888 & \textbf{5.168} & 0.991 & 0.619 & 131.7\\
 & SAM3       & 135.9 & 0.885 & 5.149 & 0.991 & \textbf{0.618} & 127.9\\
 & UniVidX$^{\dagger}$ & 143.8 & 0.866 & 5.155 & 0.992 & 0.633 & 138.5\\
\midrule
\multirow{4}{*}{CogVideoX1.5-I2V}
 & BiRefNet   & 179.5 & 0.860 & 4.868 & 0.988 & 0.640 & 128.3\\
 & MatAnyone2 & 175.2 & 0.865 & 4.876 & 0.988 & 0.621 & 125.3\\
 & SAM3       & 176.4 & 0.859 & 4.872 & 0.987 & 0.691 & 128.2\\
 & UniVidX$^{\dagger}$ & 178.2 & 0.854 & 4.870 & 0.988 & 0.652 & 128.3\\
\midrule
\multirow{4}{*}{HunyuanVideo-I2V}
 & BiRefNet   & 423.3 & 0.659 & 4.793 & 0.992 & 0.701 & 283.4\\
 & MatAnyone2 & 372.2 & 0.697 & 4.730 & 0.993 & 0.670 & 234.6\\
 & SAM3       & 353.6 & 0.719 & 4.739 & 0.991 & 0.668 & 245.0\\
 & UniVidX$^{\dagger}$ & 341.6 & 0.581 & 4.582 & 0.993 & 0.617 & 225.4\\
\midrule
\multicolumn{2}{l}{\textbf{Ours (dense)}}
 & 129.7 & 0.905 & 5.119 & \textbf{0.994} & 0.624 & \textbf{105.0}\\
\multicolumn{2}{l}{\textbf{Ours (visibility router)}}
 & \textbf{125.7} & \textbf{0.906} & 5.107 & \textbf{0.994} & 0.621 & 105.2\\
\midrule
\multicolumn{2}{l}{\emph{Ground truth (anchor)}}
 & $\approx$0 & 0.885 & 4.930 & 0.990 & 0.588 & ---\\
\bottomrule
\end{tabular}
\vspace{-1mm}
\caption{Cross-domain comparison on VideoMatte240K. Full-cohort rows use
$n{=}484$ and 97 frames; MAD uses ground-truth alpha. The separately trained
RGB-A VAE checkpoint used in this diagnostic was trained without
VideoMatte240K.
$^{\dagger}$UniVidX rows use the same seeded 400-clip manifest prefix due to
per-clip cost, so their FVD is not comparable with the $n{=}484$ rows.
DINO-ID/Aesthetic use 447--483 and 395--400 valid scores for the $n{=}484$ and
$n{=}400$ rows, respectively, after auxiliary NIQE failures; Flow-Diff uses
480--484 clips for full-cohort rows (400 for UniVidX) where foreground-flow
support is available.
\textbf{Bold} marks the best $n{=}484$ value per column, excluding the
ground-truth anchor.}
\vspace{-2mm}
\label{tab:app_ood}
\end{table*}

Table~\ref{tab:app_ood} reports the grid.
Among the directly comparable full-cohort rows, our model obtains the lowest
FVD ($125.7$ routed; $129.7$ dense) and MAD ($105.2$ routed; $105.0$ dense).
With the shared Wan backbone, generate-then-matte variants reach FVD
$132.6$--$135.9$ and MAD $127.9$--$135.9$.
These are full-system comparisons: RGB-A adaptation and routed inference vary
together, so the table does not isolate a causal gain from joint
foreground-and-alpha generation alone.
Flow Difference must still be read with content quality: Wan--SAM3 has a
slightly lower value than our router ($0.618$ vs.\ $0.621$) but a higher FVD
($135.9$ vs.\ $125.7$).

Under Stage-2 domain shift, dense fallback rises from $2.9\%$ in-domain to
$5.2\%$ ($25/484$: $22$ low-support and $3$ high-uncertainty clips).
Among clips that enter sparse execution, every clip reactivates tokens
(median $198$ tokens, versus $325$ in-domain).
Routed and dense results remain close across the reported metrics, consistent
with fallback and reactivation acting as safeguards under domain shift.


\subsection{Loss-Weighting Ablation}
\label{app:lossweight}

To test whether visibility should also reweight the training objective, we
retrain a controlled variant of the same 36-channel architecture while
changing only the spatial weighting of the flow-matching loss.
For latent cell $i$, let $a_i\in[0,1]$ denote the aligned dataset alpha
occupancy and define
\begin{equation}
  w_i=\lambda+(1-\lambda)a_i,
  \quad
  \mathcal{L}_{\lambda}
  =
  \mathbb{E}\!\left[
    \omega(t)
    \frac{\sum_i w_i
    \lVert\hat{\mathbf{v}}_{t,i}-\mathbf{v}_i\rVert_2^2}
    {16\sum_i w_i}
  \right].
  \label{eq:app_lossweight}
\end{equation}
The weight is shared by all 16 channels of the merged RGB--alpha latent.
Thus, $\lambda=1$ recovers the spatially uniform objective, whereas
$\lambda=0.2$ assigns a fully transparent cell one fifth of the weight of an
opaque cell and interpolates continuously through partial opacity.
Normalization by the mean weight keeps the overall loss scale comparable.
All data and optimization settings are unchanged.
Table~\ref{tab:app_lossweight} evaluates both models with dense inference so
that routing does not confound the comparison.

\begin{table*}[t]
  \centering
  \footnotesize
  \setlength{\tabcolsep}{4.5pt}
  \begin{tabular}{@{}lc cccccc@{}}
    \toprule
    Training objective & $\lambda$ & FVD$\downarrow$ & DINO-ID$\uparrow$ &
    Aesthetic$\uparrow$ & Motion Sm.$\uparrow$ & Flow-Diff$\downarrow$ &
    MAD\,($10^{-3}$)$\downarrow$\\
    \midrule
    Uniform & 1.0 & \textbf{172.3} & \textbf{0.840} & \textbf{5.317} &
    \textbf{0.988} & \textbf{0.550} & \textbf{108.2}\\
    Alpha-weighted & 0.2 & 185.8 & 0.836 & 5.307 &
    \textbf{0.988} & 0.560 & 108.5\\
    \bottomrule
  \end{tabular}
  \caption{\textbf{Controlled loss-weighting ablation} on the
  GameAlpha-2.4K validation set ($n{=}239$). The uniform row is the dense reference
  reported in the main paper. The alpha-weighted model has the same
  36-channel architecture and training recipe; only Eq.~\eqref{eq:app_lossweight}
  changes. Both models are evaluated with dense four-step inference.
  \textbf{Bold} marks the better value; ties at displayed precision are both
  bolded.}
  \label{tab:app_lossweight}
\end{table*}

Table~\ref{tab:app_lossweight} shows that downweighting transparent regions
raises FVD from $172.3$ to $185.8$ and
reduces DINO-ID from $0.840$ to $0.836$, while motion smoothness is unchanged
at the reported precision.
It also provides no benefit in agreement with the dataset alpha target:
Flow Difference increases from
$0.550$ to $0.560$ and MAD from $108.2$ to $108.5$.
Under this controlled setting, reallocating loss toward visible cells therefore
degrades distribution-level video quality without improving RGB--alpha
alignment or agreement with the extractor-selected alpha target, supporting
the spatially uniform objective used in the main model.
This result is specific to the tested $\lambda=0.2$ weighting and does not rule
out every alternative weighting schedule.

\subsection{Human Evaluation}
\label{sec:supp-human-evaluation}

\paragraph{Participants.}
Ten professional data annotators participated in a blinded, within-subject
evaluation. Each participant independently rated all evaluated outputs, and
participant identities are anonymized as P01--P10 in the analysis.

\paragraph{Protocol.}
We evaluate 20 reference--prompt pairs. For each pair, participants view the
checkerboard composites produced by four systems:
Wan2.1-I2V-14B (4-step) + BiRefNet,
CogVideoX1.5-5B-I2V + BiRefNet,
HunyuanVideo-I2V + BiRefNet, and our deployed visibility-routed RGBA model.
All methods are hidden behind labels A--D. Label positions are counterbalanced
over the 20 cases, such that every method appears five times at each position.
Each participant therefore rates 80 outputs on two criteria using a three-level
ordinal scale (low, medium, or high).

The two criteria displayed on the rating sheet were (1) \emph{Visual
Attractiveness}, which asks for the visual appeal of each generated
checkerboard composite, and (2) \emph{RGBA Alignment}, whether RGB and alpha
align correctly. The latter wording follows the RGBA Alignment dimension used
by TransPixeler~\cite{wang2025transpixeler}; our first criterion is not
TransPixeler's Motion Quality dimension, which instead evaluates whether
generated motion matches the corresponding text description. We therefore
report the rating-sheet construct under its literal name and do not claim to
replicate TransPixeler's evaluation dimensions or pairwise-preference protocol.
Checkerboard ratings provide a perceptual measure of visible RGBA alignment,
not pixel-accurate alpha quality or arbitrary-background recompositability.

\begin{table*}[t]
\centering
\footnotesize
\setlength{\tabcolsep}{3.4pt}
\begin{tabular}{l rrr rrr}
\toprule
& \multicolumn{3}{c}{Visual Attractiveness (\%)}
& \multicolumn{3}{c}{RGBA Alignment (\%)} \\
\cmidrule(lr){2-4}\cmidrule(lr){5-7}
Method & Low & Mid & High & Low & Mid & High \\
\midrule
Wan2.1-I2V + BiRefNet       & 12.0 & 30.0 & 58.0 & 17.5 & 28.0 & 54.5  \\
CogVideoX1.5-I2V + BiRefNet & 81.5 & 16.0 & 2.5 & 7.5 & 40.0 & 52.5  \\
HunyuanVideo-I2V + BiRefNet & 67.0 & 28.5 &  4.5 & 31.0 & 28.0 & 41.0 \\
Ours                         & 11.0 & 27.0 & \textbf{62.0} & 4.5 & 19.5 & \textbf{76.0}\\
\bottomrule
\end{tabular}
\caption{Blinded human evaluation on 20 cases by 10 professional data
annotators. Each method receives 200 ratings per criterion. Percentages sum to
100 within each method and criterion. The evaluated cases form an enriched
qualitative pool rather than a random sample of the full validation set.}
\label{tab:supp-human-evaluation}
\end{table*}

\paragraph{Analysis.}
We map low, medium, and high to 1, 2, and 3 only for descriptive summaries.
The participant is the independent unit: for each method and criterion, we
first average over the same 20 cases within each participant. We then compare
our method against each baseline using exact two-sided Wilcoxon signed-rank
tests on the 10 paired participant summaries and apply Holm correction over
the three baseline comparisons within each criterion. We additionally report
the complete ordinal-response distributions in
Table~\ref{tab:supp-human-evaluation}. Ordinal Krippendorff's alpha is .679 for
Visual Attractiveness and .508 for RGBA Alignment.

\paragraph{Results.}
For Visual Attractiveness, our method obtains the largest share of high ratings
(62.0\%), followed by Wan2.1 + BiRefNet (58.0\%). Participant-level
comparisons show that our method receives higher ratings than
CogVideoX1.5 + BiRefNet and HunyuanVideo + BiRefNet
(Holm-adjusted $p=.0059$ for both), while no significant difference is
observed relative to Wan2.1 + BiRefNet ($p=.5508$).

For RGBA Alignment, our method achieves the largest share of high ratings
(76.0\%), substantially exceeding Wan2.1 + BiRefNet (54.5\%),
CogVideoX1.5 + BiRefNet (52.5\%), and HunyuanVideo + BiRefNet
(41.0\%). Participant-level comparisons further show that our method
receives significantly higher RGBA Alignment ratings than all three
generate-then-matte baselines: Wan2.1 + BiRefNet
(Holm-adjusted $p=.0254$), CogVideoX1.5 + BiRefNet ($p=.0117$), and
HunyuanVideo + BiRefNet ($p=.0059$).

Overall, our method achieves the strongest perceptual evaluation profile:
it obtains the highest high-rating shares for both criteria, significantly
improves RGBA Alignment over all three generate-then-matte baselines,
and receives significantly higher Visual Attractiveness ratings than
CogVideoX1.5 and HunyuanVideo, while its paired comparison with Wan2.1 is not
significant.

\section{Additional Qualitative Results}
\label{app:qual}

\subsection{Full Generator--Extractor Grid}
\label{app:qual_grid}

\begin{figure*}[p]
  \centering
  \includegraphics[width=0.98\textwidth]{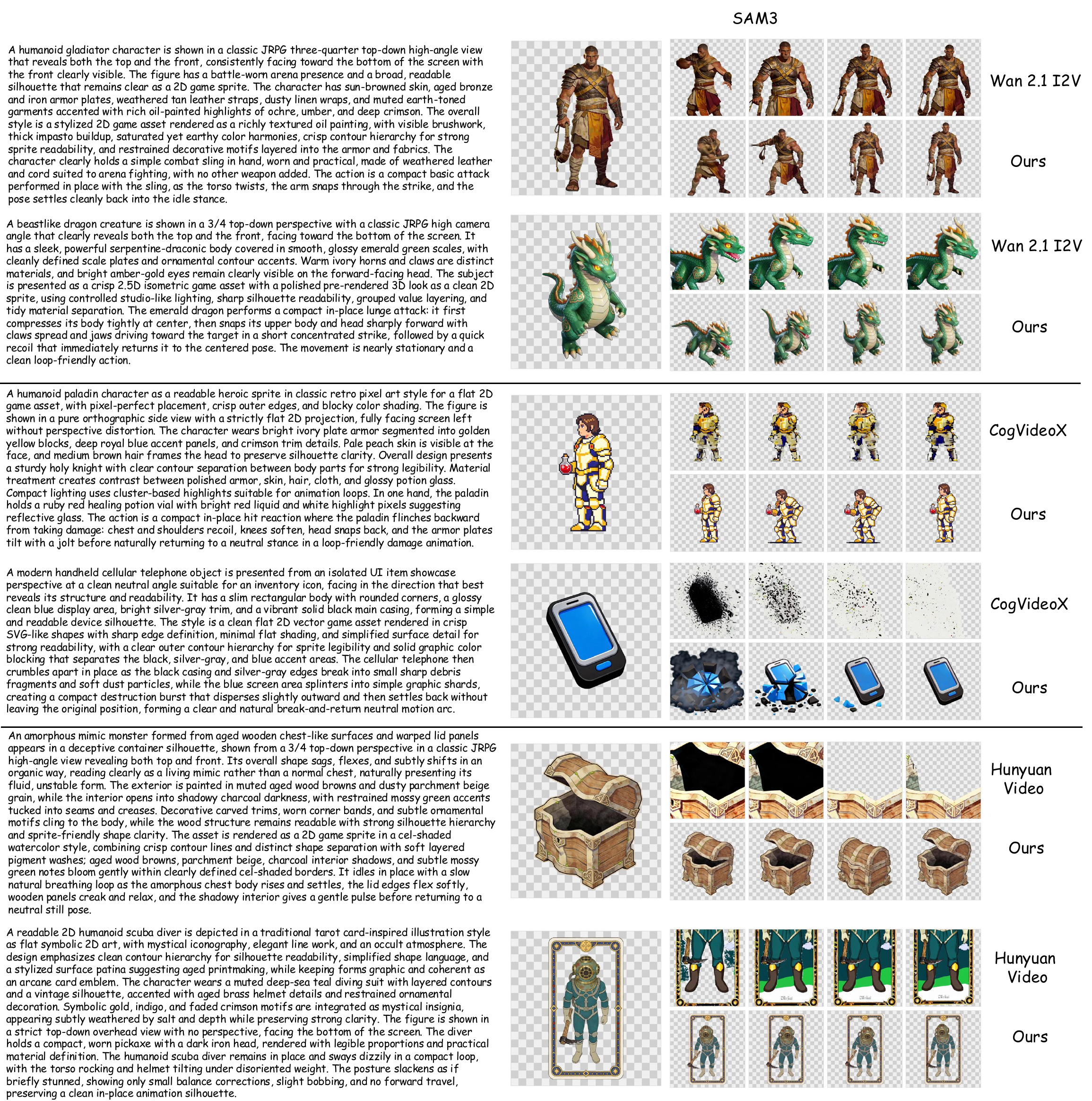}
  \caption{\textbf{Additional qualitative comparison with SAM3.}
  Each example shows the prompt (left), common RGBA reference (center), and
  four frames at the same temporal indices from the indicated RGB generator
  followed by SAM3 (top) and from our single-stage model (bottom).
  Row pairs use Wan2.1-I2V, CogVideoX1.5-I2V, and HunyuanVideo-I2V from top to
  bottom; checkerboards denote transparency.}
  \label{fig:app_qual_sam3}
\end{figure*}

\begin{figure*}[p]
  \centering
  \includegraphics[width=0.98\textwidth]{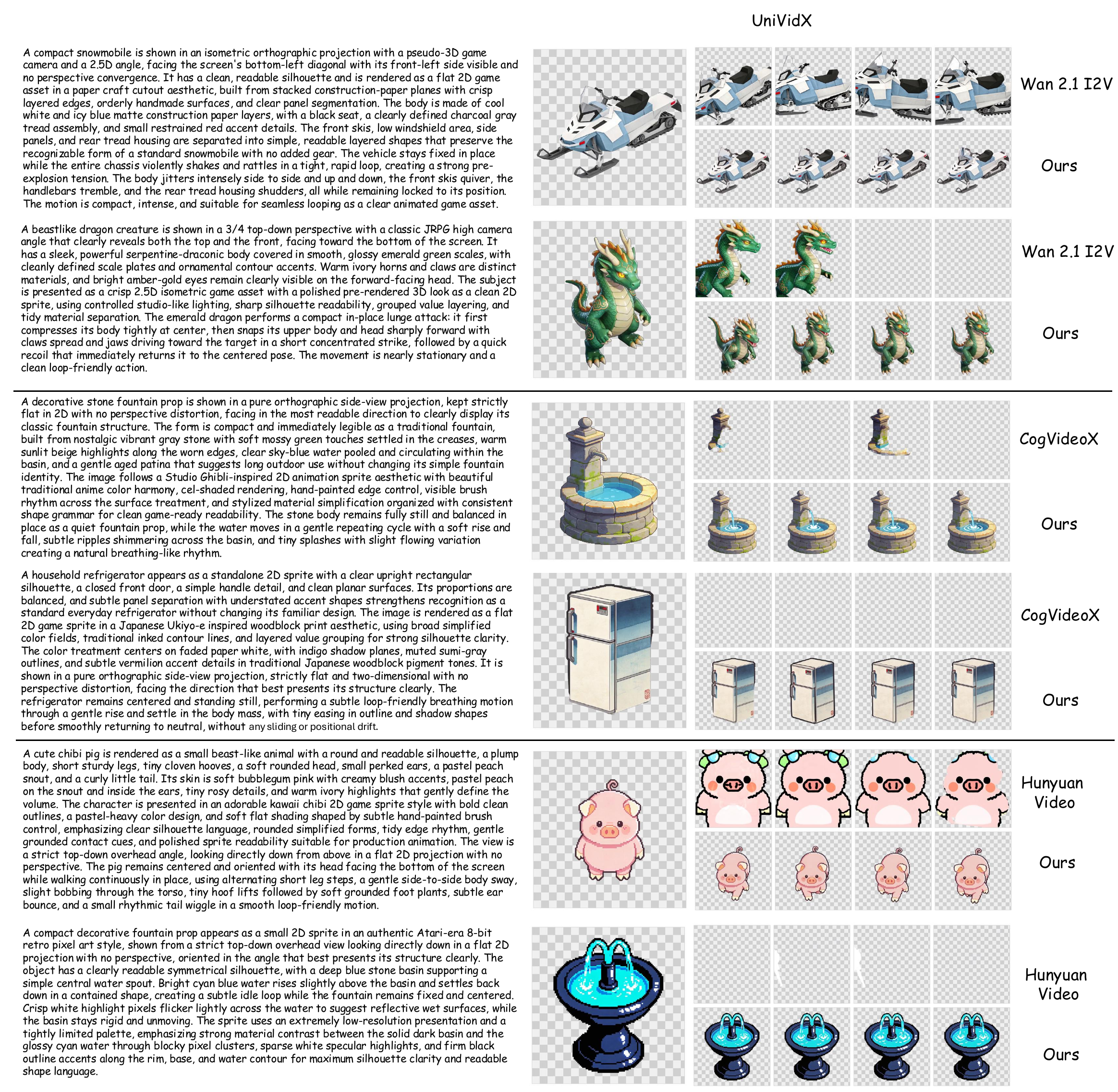}
  \caption{\textbf{Additional qualitative comparison with UniVidX.}
  The layout follows Figure~\ref{fig:app_qual_sam3}: the upper and lower rows
  show the same four temporal indices from the indicated RGB generator
  followed by UniVidX and from our model, respectively, under the same prompt
  and reference.
  Checkerboards denote transparency; checkerboard-only panels contain no
  visible foreground in the displayed RGBA output.}
  \label{fig:app_qual_unividx}
\end{figure*}

Together with the qualitative comparison in the main paper,
Figures~\ref{fig:app_qual_sam3} and~\ref{fig:app_qual_unividx} cover the
three RGB generators and four post-hoc extractors in the quantitative grid.
Within each example, the paired rows use the same reference and prompt.

\subsection{Qualitative Routing Ablation}
\label{app:qual_ablation}

\begin{figure*}[p]
  \centering
  \includegraphics[width=\textwidth,trim=0 773bp 0 0,clip]{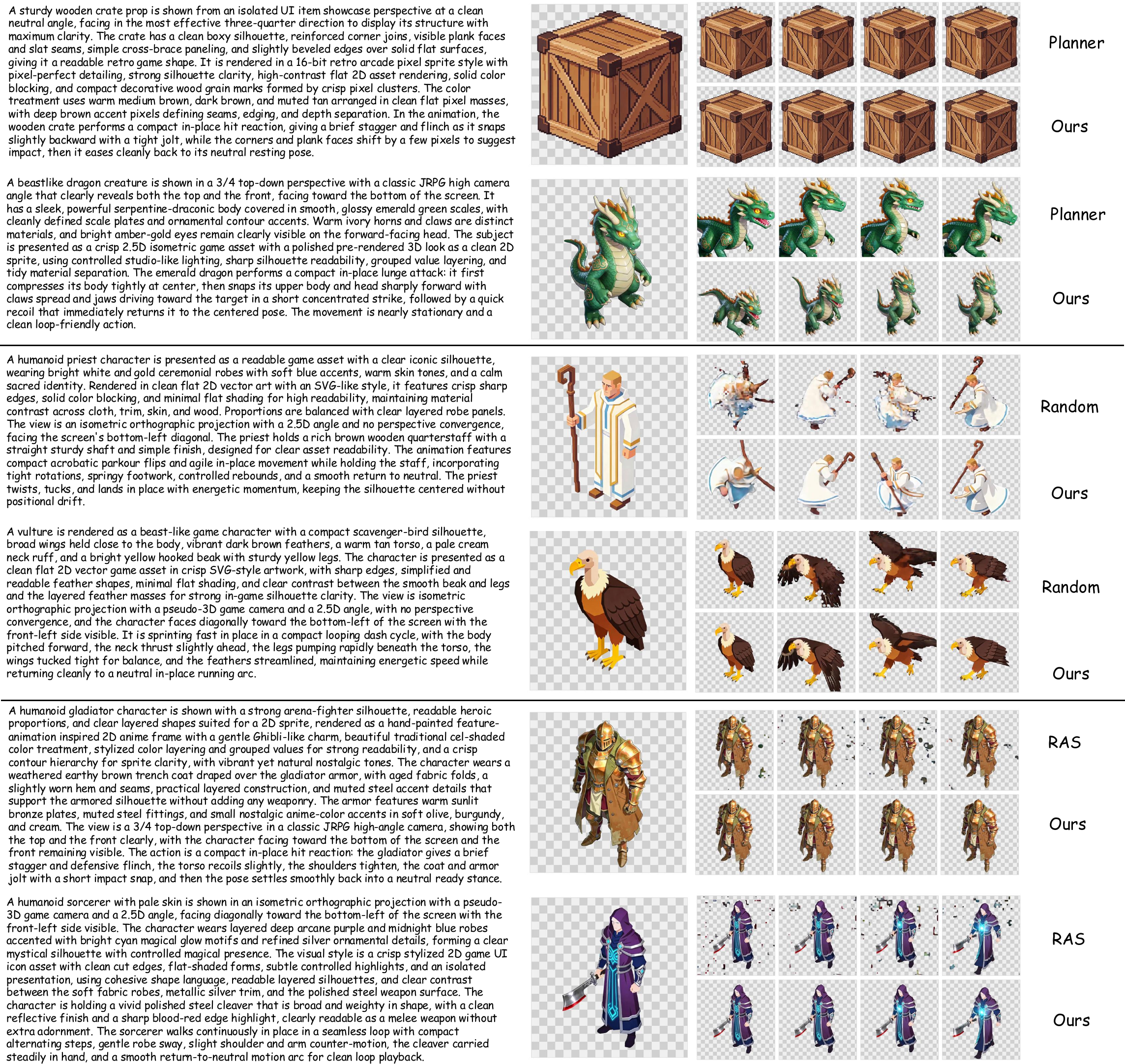}
  \caption{\textbf{Qualitative routing ablation: selection policies.}
  Each example shows the prompt (left), common RGBA reference (center), and
  the same four temporal indices from the named alternative (top) and our
  full visibility router (bottom).
  Two cases each compare Planner, Random, and RAS under the matched
  realized-skip protocol reported in the main paper.
  Checkerboards denote transparency.}
  \label{fig:app_qual_routing_signal}
\end{figure*}

Figures~\ref{fig:app_qual_routing_signal}
and~\ref{fig:app_qual_routing_safety} complement the quantitative routing
ablation in the main paper with selected paired examples.

\begin{figure*}[p]
  \centering
  \includegraphics[width=\textwidth,trim=0 0 0 1038bp,clip]{figure/sup_qual_ablation.pdf}
  \caption{\textbf{Qualitative routing ablation: static support and
  reactivation.}
  The layout follows Figure~\ref{fig:app_qual_routing_signal}.
  Union2D follows the same matched realized-skip protocol, whereas the
  no-reactivation variant shares our evaluation-2 routing decision but fixes
  the routed set through evaluations~3--4.
  The selected cases visualize how static support and local revision affect
  foreground structure over time.}
  \label{fig:app_qual_routing_safety}
\end{figure*}

\subsection{Prompts of the Qualitative Examples}
\label{app:prompts}

The main paper's qualitative comparison pairs three RGB generators with
BiRefNet or MatAnyone2.
Within each group, our model and the corresponding two-stage pipeline receive
the identical text prompt reproduced below and the identical
white-composited reference frame; the prompts are quoted verbatim, and the
sampled entity, equipment, action, style, and viewpoint of
Table~\ref{tab:app_taxonomy} remain traceable in each of them.

{\small
\noindent\textbf{Top row, left --- Wan2.1-I2V (4-step) $+$ BiRefNet.}
A rugged humanoid barbarian with weathered tan skin, a sturdy survivalist
silhouette, and a battle-worn presence --- grounded proportions and practical
anatomy, a believable fantasy warrior rather than an exaggerated monster.
Rendered as hand-drawn 2D game concept art with clean readable linework,
controlled contour hierarchy, and painterly digital rendering, using
expressive brushwork and stylized textures to define materials clearly while
preserving strong sprite legibility and cohesive shape language. Strict
top-down overhead view looking directly downward in flat 2D projection with
no perspective, the character facing the bottom of the screen. He wears a
worn mask of carved bone or aged wood, hand-shaped with natural
brush-painted markings and a scuffed surface. The outfit is rugged
earth-toned leather with muted iron and bronze accents --- functional,
distressed, repeatedly repaired, no weapon. He continuously climbs a vertical
rope in place in a seamless loop, alternating hand-over-hand pulls with
compact leg engagement and controlled body lift while staying centered. The
motion is grounded and rhythmic, with subtle torso compression and extension,
slight movement in the leathers and straps, and the mask stable as he ascends
without changing screen position.

\smallskip
\noindent\textbf{Top row, right --- Wan2.1-I2V (4-step) $+$ MatAnyone2.}
A humanoid vampire creature is shown in a strict top-down overhead view, seen
directly from above in a flat 2D projection with no perspective, oriented
toward screen left. It has pale, ashen skin, sharp predatory facial features,
and deep crimson accents integrated into its design. The body is lean and
undead-looking, with a clear readable silhouette and gothic menace. Layered
shadowy black and burgundy garments frame the form, emphasizing clawed hands,
tense posture, and a cold nocturnal presence. Over this clothing it wears a
worn dark bulletproof tactical vest fitted close to the body, scuffed,
utilitarian, and slightly weathered, rendered in natural hand-painted tones
with stylized digital brush texture while remaining clearly protective modern
armor rather than decorative costume. The image is rendered as hand-drawn 2D
game concept art with clean controlled linework, painterly digital rendering,
and expressive brush-driven surface treatment, using strong silhouette
readability, cohesive shape language, and stylized but natural material
definition suited for a polished 2D game asset, with crisp forms, selective
texture, and a cinematic yet objective presentation. The humanoid vampire
performs a compact vertical jump in place, springing upward and landing back
in the exact same spot with a clean return to a neutral pose. During the lift
and descent, the garments and the worn dark tactical vest react subtly to the
motion while the body remains centered, the silhouette stays stable and
readable, and the landing resolves into a clear idle-ready stance.

\smallskip
\noindent\textbf{Middle row, left --- CogVideoX1.5-I2V $+$ BiRefNet.}
A cute chibi oni demon creature is shown in a strict top-down overhead view,
looking directly down from above in a flat 2D projection with no perspective,
facing the top of the screen. It has a compact humanoid body, an oversized
head, a tiny torso and limbs, and a clean readable silhouette. Its skin is
pastel cherry red, with small baby pink horns, creamy ivory fangs peeking
from a playful mouth, round bubblegum pink cheeks, and soft lavender shadow
accents defining simple stylized forms. The visual treatment is a kawaii
chibi 2D sprite style with thick crisp outlines, a pastel candy palette, soft
flat shading, minimal but polished surface detail, and strong shape-language
consistency for clear game-asset silhouette readability. It is holding a
candy-colored crossbow sized for the chibi character, with a chunky readable
shape, a mint green body, baby blue limbs and grip details, and pale yellow
accent pieces, built with simplified 2D game-friendly construction, soft flat
color blocks, and bold outline clarity. The action is an in-place shooting
animation: the chibi oni plants its stance, raises the mint-and-blue crossbow
to aim, gives a compact firing snap with a small recoil through the arms and
shoulders, and then smoothly returns along a clean arc to a neutral ready
pose.

\smallskip
\noindent\textbf{Middle row, right --- CogVideoX1.5-I2V $+$ MatAnyone2.}
A gargoyle creature with a beast-like silhouette is shown in an isometric
orthographic projection with a pseudo-3D game camera at a 2.5D angle, with no
perspective convergence, facing diagonally toward the bottom-right of the
screen with its front-right side clearly visible. It has carved-stone anatomy
and rugged architectural forms, broad clawed limbs, heavy wings, a horned
head, and a snarling face. Its surface is weathered stone gray with cool blue
undertones, broken by mossy olive staining, soot-dark recessed cracks,
chipped edges, and muted hand-painted earthy accents. Strong material
contrast appears across rough rock plates, worn ornamented ridges, and subtle
decorative carved motifs embedded in the body. The image is rendered as
hand-drawn 2D game concept art with crisp clean linework, painterly digital
rendering, and expressive brush-driven strokes, designed as a readable 2D
game asset with a clear silhouette. The gargoyle remains rooted in place,
swaying dizzily from side to side in a loose off-balance posture, with its
head wobbling, shoulders slumped, and wings twitching unevenly. Its clawed
hands hang open and empty, and the beast-like stone body rocks in a compact
loop that clearly conveys a stunned status effect without any forward
movement.

\smallskip
\noindent\textbf{Bottom row, left --- HunyuanVideo-I2V $+$ BiRefNet.}
An amorphous air elemental creature formed from swirling sky-blue, white, and
pale cyan vapor masses appears as a readable elemental body silhouette built
from soft cloudlike contours, curling wind ribbons, and floating wisps,
accented with bright solid vector highlights and subtle ornamental airflow
motifs. The subject is rendered as a clean 2D flat vector game asset in crisp
SVG-inspired styling, with sharp polished edges, minimal flat shading,
clearly grouped color planes, and a palette of bright sky blue, white, and
pale cyan enhanced by vibrant solid highlights for a stylized yet
production-ready look. The view remains a strict top-down overhead angle,
looking directly down from above in a 2D flat projection with no perspective,
and the creature has no specific facing so its fluid amorphous form is shown
naturally. In the looping animation, the air elemental stays in place while
steadily dripping and leaking fluid beneath itself, with small repeated
droplets forming along its lower vapor edges and falling straight downward in
a compact cycle, as the main body gently quivers and ripples without shifting
position.

\smallskip
\noindent\textbf{Bottom row, right --- HunyuanVideo-I2V $+$ MatAnyone2.}
A sturdy wooden crate prop is shown from an isolated UI item showcase
perspective at a clean neutral angle, facing in the most effective
three-quarter direction to display its structure with maximum clarity. The
crate has a clean boxy silhouette, reinforced corner joins, visible plank
faces and slat seams, simple cross-brace paneling, and slightly beveled edges
over solid flat surfaces, giving it a readable retro game shape. It is
rendered in a 16-bit retro arcade pixel sprite style with pixel-perfect
detailing, strong silhouette clarity, high-contrast flat 2D asset rendering,
solid color blocking, and compact decorative wood grain marks formed by crisp
pixel clusters. The color treatment uses warm medium brown, dark brown, and
muted tan arranged in clean flat pixel masses, with deep brown accent pixels
defining seams, edging, and depth separation. In the animation, the wooden
crate performs a compact in-place hit reaction, giving a brief stagger and
flinch as it snaps slightly backward with a tight jolt, while the corners and
plank faces shift by a few pixels to suggest impact, then it eases cleanly
back to its neutral resting pose.
}

\section{Reproducibility Notes}
\label{app:repro}

GameAlpha-2.4K is split 9:1 before any training; the Stage-1 VAE trains on the
training split plus VideoMatte240K, the Stage-2 DiT only on the training
split, and the transparency probe is fitted on dense trajectories of training
clips and frozen.
The cross-domain diagnostic in Appendix~\ref{app:ood} is a stated exception:
it uses a separately trained RGB-A VAE checkpoint that did not use
VideoMatte240K.
Unless otherwise stated, the main-paper GameAlpha automatic metrics use the
full validation split ($n=239$) at a fixed seed; auxiliary analyses,
cross-domain experiments, and human evaluation use the sample sizes stated in
their respective text or captions. The analysis that selects
$k_{\mathrm{route}}=2$ and $r_s=2$ is described in the main paper.
Further implementation details are intended for the planned dataset, code,
and checkpoint releases; release status and licensing are described in
Appendix~\ref{app:licenses}.

\section{Licenses and Intended Release}
\label{app:licenses}

GameAlpha-2.4K and our implementation are not public at submission.
We plan to release GameAlpha-2.4K, open-source the accompanying implementation,
and release the trained model checkpoint. The artifact scope and applicable
licenses will be specified at release.

\paragraph{Verified third-party terms.}
At submission, the official release pages report the following licenses;
repository code, model weights, and datasets are distinguished because one
artifact's license does not automatically cover the others.
\begin{itemize}
  \item \textbf{Apache License 2.0.}
  Wan2.1 code and released models~\cite{wan2025}; LightX2V code and the
  four-step I2V distillation checkpoint used here~\cite{lightx2v}; standard
  DINOv2 code and weights~\cite{oquab2024dinov2}; VBench code~\cite{huang2023vbench};
  and UniVidX code and released checkpoints~\cite{chen2026unividx}.

  \item \textbf{MIT License.}
  The BiRefNet~\cite{zheng2024birefnet} and Wan-Alpha~\cite{dong2025wanalpha}
  repository code and the LAION aesthetic-predictor repository use MIT.
  Wan-Alpha's released model cards separately label its weights Apache-2.0.

  \item \textbf{Custom or restricted licenses.}
  CogVideo code is Apache-2.0, but the CogVideoX1.5-5B-I2V parameters use the
  CogVideoX License~\cite{yang2024cogvideox}.
  HunyuanVideo-I2V code and weights use the Tencent Hunyuan Community License
  Agreement~\cite{kong2024hunyuanvideo}.
  MatAnyone2 uses S-Lab License 1.0, which permits non-commercial use and
  requires permission for commercial use~\cite{yang2026matanyone2}; SAM3 code
  and weights use the SAM License~\cite{carion2025sam3segmentconcepts}; and the
  RMBG-2.0 weights used with the BiRefNet architecture use CC BY-NC 4.0, with
  commercial use requiring a Bria agreement.

  \item \textbf{Data and metric caveats.}
  The official VideoMatte240K page~\cite{lin2021real} does not specify a
  separate dataset license; we do not treat the associated repository's MIT
  software license as licensing the dataset.
  The StyleGAN-V I3D/FVD implementation used for FVD~\cite{skorokhodov2021stylegan} has no
  standalone license file; its repository states that it is likely subject to
  the NVIDIA Source Code License-NC inherited through StyleGAN2-ADA.
\end{itemize}

All third-party resources, including bundled submodels and dependencies,
remain governed by their upstream terms and are not relicensed by us.
The planned release will reference upstream sources and record artifact-level
licenses rather than redistribute third-party datasets or checkpoints unless
their terms permit it.
The fixed generation and screening services used for dataset construction are
governed by their providers' service terms.
Before releasing GameAlpha-2.4K, we will verify that those terms permit
redistribution of the generated outputs.

\bibliography{aaai2027}

\end{document}